\documentclass[10pt,twocolumn,letterpaper]{article}

\usepackage{cvpr}              
\usepackage{newtxtext,newtxmath} 
\usepackage[T1]{fontenc}        
\def\eqref#1{equation~\ref{#1}}

\def\1{\bm{1}}

\def\vmu{{\bm{\mu}}}

\def\vc{{\bm{c}}}

\def\vf{{\bm{f}}}
\def\vg{{\bm{g}}}
\def\vh{{\bm{h}}}

\def\vl{{\bm{l}}}
\def\vm{{\bm{m}}}

\def\vp{{\bm{p}}}
\def\vq{{\bm{q}}}

\def\vs{{\bm{s}}}

\def\vx{{\bm{x}}}

\def\mA{{\bm{A}}}

\def\mG{{\bm{G}}}
\def\mH{{\bm{H}}}

\def\mL{{\bm{L}}}

\def\sR{{\mathbb{R}}}

\usepackage[acronym]{glossaries}
\usepackage[table, dvipsnames]{xcolor}
\usepackage{bm}
\usepackage{multirow} 

\newacronym{3dgs}{3DGS}{3D Gaussian Splatting}
\newacronym{superg}{SuperG}{Super-Gaussian}
\newacronym{mIoU}{mIoU}{mean Intersection over Union}
\newacronym{mAcc}{mAcc}{mean Accuracy}

\newcommand\blfootnote[1]{%
  \begingroup
  \renewcommand\thefootnote{}\footnote{#1}%
  \addtocounter{footnote}{-1}%
  \endgroup
}

\definecolor{cvprblue}{rgb}{0.21,0.49,0.74}
\usepackage[pagebackref,breaklinks,colorlinks,citecolor=cvprblue]{hyperref}
\renewcommand{\sectionautorefname}{Section}

\title{SuperGSeg: Open-Vocabulary 3D Segmentation with\\Structured Super-Gaussians}

\author{
    Siyun Liang$^{1,2 *}$\blfootnote{1}\quad
    Sen Wang$^{1,4 *}$\blfootnote{1}\quad
    Kunyi Li$^{1,4}$ \quad
    Michael Niemeyer$^{3}$ \quad
    Stefano Gasperini$^{1,4,5}$ \\
    Hendrik P.A. Lensch$^{2}$ \quad
    Nassir Navab$^{1}$ \quad 
    Federico Tombari$^{1,3}$ \\
    {\normalsize $^{1}$Technical University of Munich} \quad
    {\normalsize $^{2}$University of Tübingen} \quad \\
    {\normalsize $^{3}$Google} \quad 
    {\normalsize $^{4}$Munich Center for Machine Learning} \quad 
    {\normalsize $^{5}$VisualAIs} \\
}

\begin{document}
\crefname{section}{Section}{Sections}
\setlength{\abovedisplayskip}{3pt}
\setlength{\belowdisplayskip}{3pt}
\setlength{\abovedisplayshortskip}{2pt}
\setlength{\belowdisplayshortskip}{2pt}

\twocolumn[{%
    \renewcommand\twocolumn[1][]{#1}
    \maketitle
    \thispagestyle{empty}
    
    \vspace{-8mm}
    
    \begin{center}
        \includegraphics[width=0.95\linewidth]{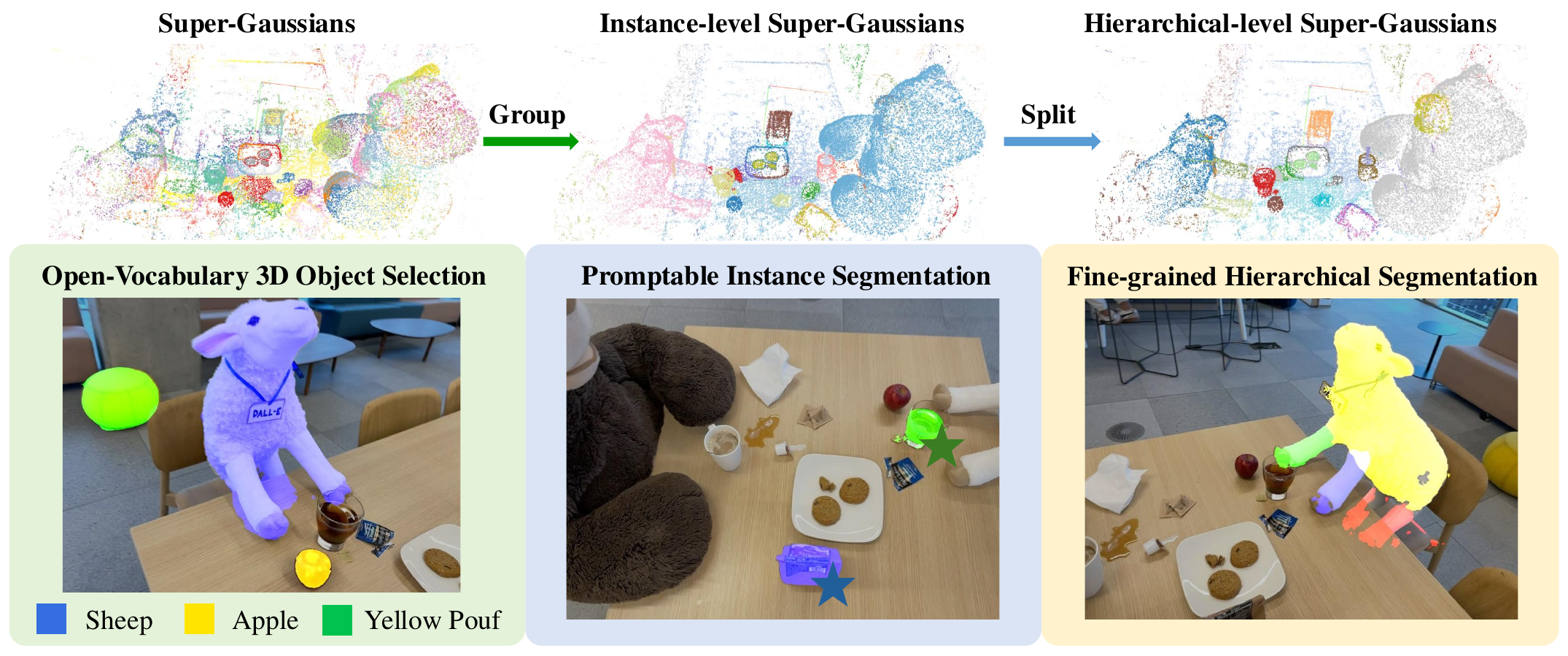}
        \vspace{-2mm}
        \captionof{figure}{We present \textbf{SuperGSeg}, a novel method that clusters similar Gaussians into superpoint-like representations, termed \acrlong{superg}s~(\acrshort{superg}s). SuperGSeg enables efficient integration of diverse feature fields for comprehensive 3D scene understanding. \textcolor[RGB]{139,182,101}{Left}: Querying \acrshort{superg}s' language features enables open-vocabulary 3D object selection, producing consistent 3D masks that extend beyond 2D visible surfaces, e.g., the leg of the sheep under the table. \textcolor[RGB]{97,128,191}{Middle}: Grouping \acrshort{superg}s by instance features enables promptable instance segmentation. \textcolor[RGB]{243,205,85}{Right}: Further splitting instances via hierarchical features enables fine-grained hierarchical segmentation.}
        \label{fig:teaser}
    \end{center}
}]

\blfootnote{* Equal contribution.}

\nolinenumbers
\begin{abstract}

3D Gaussian Splatting has recently gained traction for its efficient training and real-time rendering. While its vanilla representation is mainly designed for view synthesis, recent works extended it to scene understanding with language features.
However, storing additional high-dimensional features per Gaussian for semantic information is memory-intensive, which limits their ability to segment and interpret challenging scenes. 
To this end, we introduce SuperGSeg, a novel approach that fosters cohesive, context-aware hierarchical scene representation by disentangling segmentation and language field distillation.
SuperGSeg first employs neural 3D Gaussians to learn geometry, instance and hierarchical segmentation features from multi-view images with the aid of off-the-shelf 2D masks. 
These features are then leveraged to create a sparse set of \acrlong{superg}s. \acrlong{superg}s facilitate the lifting and distillation of 2D language features into 3D space. They enable hierarchical scene understanding with high-dimensional language feature rendering at moderate GPU memory costs. Extensive experiments demonstrate that SuperGSeg achieves remarkable performance on both open-vocabulary object selection and semantic segmentation tasks. More results at \url{supergseg.github.io}.
\end{abstract}    
\section{Introduction}
\label{sec:intro}

\acrlong{3dgs}~(\acrshort{3dgs})~\cite{kerbl20233d} has rapidly emerged as a compelling alternative to NeRF~\cite{mildenhall2020nerf} for its efficient training, real-time rendering, and explicit 3D representation. These advantages make \acrshort{3dgs} well-suited for a broad range of applications, including 3D reconstruction~\cite{yu2024gaussian, dai2024high, guedon2024sugar}, content generation~\cite{yi2024gaussiandreamerpro}, and scene understanding~\cite{qin2024langsplat, shi2024language, zhou2024feature, gaussian_grouping, wu2024opengaussian, li2024instancegaussian}. A particularly promising direction involves extending \acrshort{3dgs} frameworks to open-vocabulary understanding, enabling flexible, language-driven interaction with 3D scenes~\cite{radford2021clip, li2022languagedriven}.

Several recent methods aim to enable such open-vocabulary capabilities in \acrshort{3dgs} by distilling language features from both 2D~\cite{zuo2024fmgs, qin2024langsplat, li2024geogaussian, zhou2024feature} and 3D~\cite{wu2024opengaussian, li2024instancegaussian} perspectives. In 2D-based methods, language features extracted from images are lifted into 3D by exploiting the multi-view consistency inherent in \acrshort{3dgs} rendering. To reduce the substantial memory and computation overhead of storing and processing high-dimensional language features for each Gaussian, these methods employ dimensionality reduction techniques~\cite{qin2024langsplat, zhou2024feature}. However, this compression inevitably discards fine-grained semantic information. Another limitation is their inability to recognize partially occluded objects, which is often necessary in 3D understanding tasks. Text queries are performed on rendered pixels, which only capture the visible surface along each viewing ray. Consequently, objects that are partially or fully hidden cannot be retrieved. In contrast, 3D methods~\cite{wu2024opengaussian, li2024instancegaussian} perform text queries directly in 3D space at the point level, which enables the retrieval of occluded objects by rendering the queried Gaussians into masks (see \figureautorefname~\ref{fig:lerf_qualitative}), but also introduces new limitations. By directly associating language features with individual Gaussians and decoupling alpha blending, they cannot render consistent language feature maps in pixel space, which in turn makes them unsuitable for tasks such as pixel-wise dense semantic segmentation in 2D.

To address the aforementioned issues, we introduce a novel approach that: (1) preserves high-dimensional language feature embeddings without information loss, (2) handles occlusions by operating directly in 3D space, and (3) supports multi-granular segmentation, ultimately enabling open-vocabulary queries in both 2D and 3D, as shown in \figureautorefname~\ref{fig:teaser}. Inspired by superpoints~\cite{nguyen2024open3dis} in point cloud analysis, our method clusters millions of Gaussians into a compact set of \acrlong{superg}s~(\acrshort{superg}s). However, due to the inherent noise in Gaussian point clouds, clustering solely based on Gaussian positions often produces suboptimal groupings. Instead, we leverage instance and hierarchical features extracted from grouped SAM masks~\cite{ying2024omniseg3d} to guide clustering via an adaptive online clustering network~\cite{hui2021spnet}. For open-vocabulary scene understanding, we further distill 2D CLIP features~\cite{radford2021clip} onto \acrshort{superg}s that integrate both spatial and semantic information. This compact representation allows language features to be assigned at the \acrshort{superg} level rather than to each individual Gaussian~\cite{shi2024language, zhou2024feature, qin2024langsplat}, thereby reducing the number of learnable language features from millions to only thousands, significantly lowering memory usage while retaining the full descriptive power of the original high-dimensional features.

Extensive experiments on the LERF-OVS~\cite{qin2024langsplat} and ScanNet~\cite{dai2017scannet} datasets show that our method achieves remarkable performance in open-vocabulary 3D object retrieval and scene-level semantic segmentation, demonstrating superior capability in producing complete and consistent masks for 3D object retrieval and capturing fine-grained scene details for 2D dense pixel-wise segmentation. We summarize the main contributions as follows:
\begin{itemize}
    \item We introduce SuperGSeg, a novel 3D scene understanding framework built on \acrlong{superg} representations, enabling effective high-dimensional language feature distillation without information loss.
    \item We propose a novel neural Gaussian rasterization pipeline that distills instance and hierarchical feature fields, facilitating \acrlong{superg} clustering and supporting multi-granular scene understanding.
    \item We design an online clustering network that adaptively fuses geometric, semantic, and appearance cues to generate \acrlong{superg}s, thus improving clustering quality.
\end{itemize}
\section{Related Work}
\label{sec:literature}

\begin{figure*}
    \centering 
    \includegraphics[width=\linewidth]{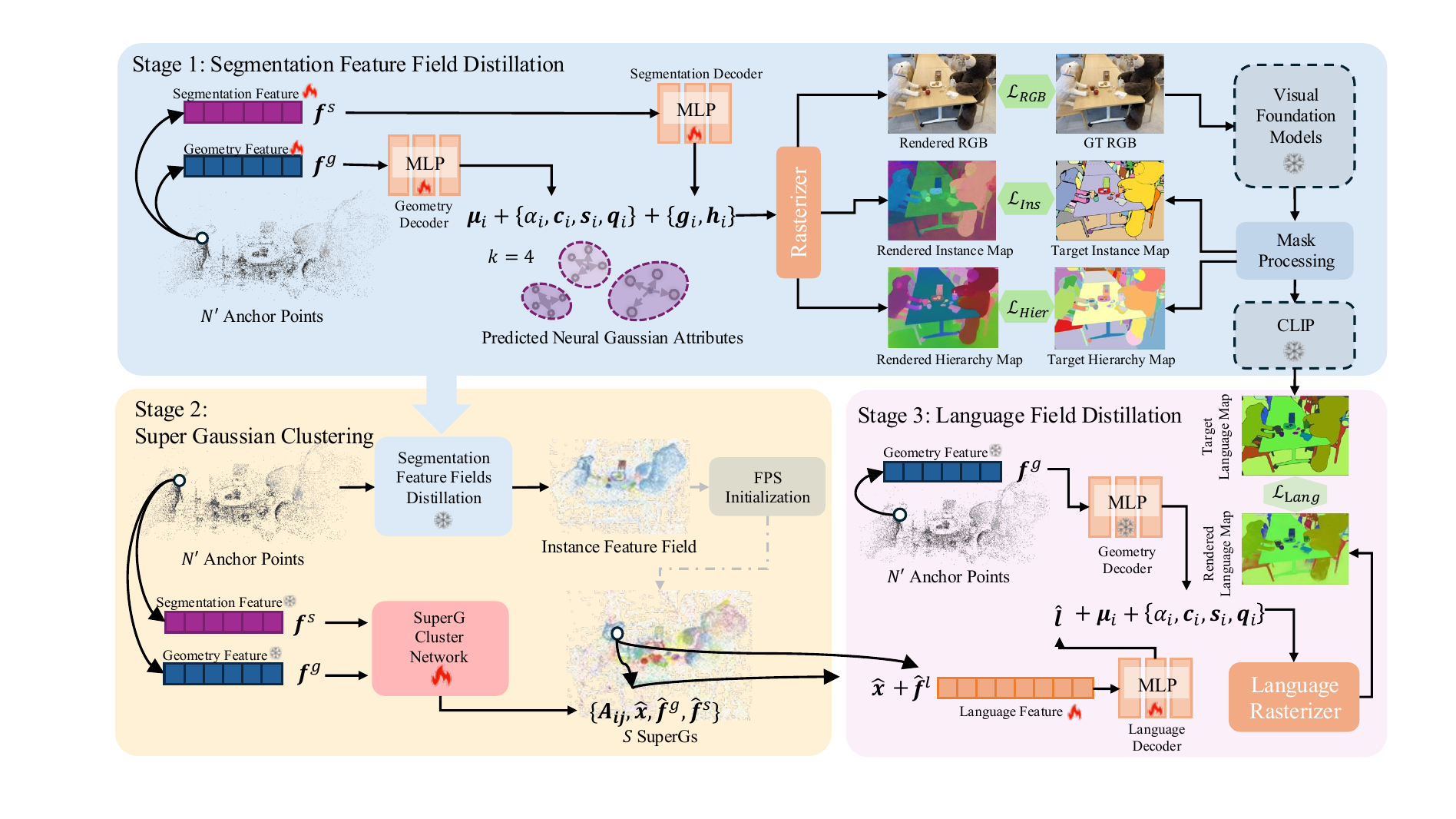} 
    \vspace{-12mm}
    \caption{\textbf{SuperGSeg Overview.} 
    We initialize the 3D Gaussians from a sparse set of anchor points, each generating \(k\) Gaussians with corresponding attributes. First, we train the appearance and segmentation features using RGB images and segmentation masks generated by SAM~\cite{kirillov2023sam}. Next, we use the segmentation features and their spatial positions to produce a sparse set of \acrlong{superg}s, each carrying a 512-dimensional language feature. Finally, we train this high-dimensional language feature using a 2D feature map from CLIP~\cite{radford2021clip}.}
    \label{fig:overview}
    \vspace{-0.2cm}
\end{figure*}

\textbf{3D Open-Vocabulary Understanding.} Advancements in universal 2D scene understanding, driven by foundation models such as CLIP~\cite{radford2021clip} and SAM~\cite{kirillov2023sam}, have motivated the integration of language-aligned features into 3D scene representations. Early efforts incorporated these 2D features~\cite{radford2021clip, caron2021dino} into NeRF-based representations~\cite{kerr2023lerf, engelmann2024opennerf}, enabling open-vocabulary queries in 3D scenes but at the cost of slow rendering and high memory usage. More recently, the emergence of \acrshort{3dgs} as a high-quality, real-time alternative for novel view synthesis has inspired extensions toward 3D scene understanding. For example, LangSplat~\cite{qin2024langsplat} employs a scene-specific language autoencoder to compress high-dimensional CLIP features, providing clear object boundaries in rendered feature images while reducing memory usage. Feature3DGS~\cite{zhou2024feature} introduces a parallel Gaussian rasterizer with a lightweight convolutional decoder to distill high-dimensional features for tasks like scene editing and segmentation. However, these dimensionality reduction techniques inevitably discard fine-grained semantic information. OpenGaussian~\cite{wu2024opengaussian} instead directly associates uncompressed, lossless CLIP features with 3D Gaussians, preserving complete semantics and enabling the retrieval of visually occluded objects by performing queries directly in 3D space. Nevertheless, its decoupled language codebook design makes per-pixel 2D language feature rendering infeasible, thereby limiting performance on dense, pixel-wise semantic prediction tasks.

Despite notable progress, most existing methods focus primarily on instance-level knowledge while neglecting fine-grained part-level semantics~\cite{gaussian_grouping, wu2024opengaussian}, or require separate models for different semantic granularities~\cite{qin2024langsplat}. While recent methods~\cite{ying2024omniseg3d, kim2024garfield} explore hierarchical 3D understanding at the part-level, they lack support for open-vocabulary language queries, leaving the joint modeling of multi-granularity 3D representation with language feature largely unexplored. In contrast, our method integrates both instance and hierarchical features from 2D inputs, and introduces a \acrlong{superg} based language field that fuses segmentation information with the spatial distribution of 3D Gaussians, thereby enabling open-vocabulary, multi-granularity, and occlusion-robust 3D segmentation.

\textbf{Superpoints.}
Superpoints have long served as fundamental primitives for various point cloud understanding tasks~\cite{hui2021spnet, landrieu2018large, robert2023efficient, robert2024scalable, landrieu2017cut, zhu2024spgroup3d, cheng2021sspc}. Early approaches, such as Voxel Cloud Connectivity Segmentation (VCCS)~\cite{papon2013voxel}, segment a voxelized 3D grid into spatially coherent regions using region-growing variants of $K$-means clustering. More recent works leverage learned point cloud representations~\cite{qi2017pointnet, vaswani2017attention} to infer superpoints directly from 3D scans~\cite{landrieu2019point, hui2021spnet, robert2023efficient}. Superpoints have also been adopted for open-vocabulary 3D segmentation~\cite{nguyen2024open3dis}, demonstrating robustness in complex scenes. However, directly applying superpoint methods to \acrshort{3dgs} is challenging due to noisy Gaussian geometry. To address this, we leverage instance- and part-level cues from 2D foundation models to guide superpoint formation, effectively bridging high-quality 2D features with noisy 3D Gaussian representations.

\vspace{-1mm}
\section{Method}
\label{secs:method}

Given a set of posed RGB images, our goal is to reconstruct a 3D scene with a compact language feature field that supports open-vocabulary querying of arbitrary concepts. To achieve this, we propose a three-stage training paradigm, as shown in \figureautorefname~\ref{fig:overview}. In the first stage, we train a neural variant of 3DGS~\cite{lu2024scaffold} to reconstruct scene geometry using $N'$ anchor points, each having a geometry feature $\vf^g$ and a segmentation feature $\vf^s$. Anchor points are then spawned into a set of neural Gaussians and optimized. In the second stage, a learnable cluster network groups the anchors into $S$ \acrshort{superg}s using $\vf^g$, $\vf^s$, and anchor position $\vx$, ensuring geometric and semantic consistency. Since $S \ll N'$, this yields a far more compact representation. In the third stage, we learn a language feature $\widehat{\vf^l}$ for each \acrshort{superg}, enabling open-vocabulary queries on just $S$ \acrshort{superg}s rather than millions of individual Gaussians.

\subsection{Preliminaries: Neural Gaussian Splatting}
\label{sec:pre}
We begin with Stage 1 of our pipeline: modeling the scene geometry with Scaffold-GS~\cite{lu2024scaffold} structure. Vanilla 3DGS represents a scene with $N$ Gaussians, each parameterized by a center $\vmu$, opacity $\alpha$, color $\vc$, scale $\vs$ and quaternion $\vq$. These Gaussians are projected onto the image plane~\cite{zwicker2002ewa} and rendered into RGB images via $\alpha$-blending. While achieving leading rendering quality and speed, optimizing each Gaussian independently often leads to overfitting, redundancy, and degraded robustness in challenging regions such as texture-less surfaces. Scaffold-GS addresses these issues by voxelizing the scene into $N'$ anchor points, each at position $\vx$. From each anchor, $k$ neural Gaussians are derived, where centers are computed as $\vx$ plus learnable offsets, and the remaining attributes $(\alpha, \vc, \vs, \vq)$ are produced on the fly from the anchor’s geometry feature $\vf^g$ via dedicated MLPs. By tying Gaussians to anchors, Scaffold-GS constrains their spatial distribution to the scene structure, preventing uncontrolled growth and improving robustness.

Training in \acrshort{3dgs} typically relies on a photometric loss $\mathcal{L}_{RGB}$, where rendered RGB images are supervised against ground-truth views. Unlike vanilla 3DGS that optimizes $(\vmu, \alpha, \vc, \vs, \vq)_N$, with $N$ often reaching millions for complex scenes, Scaffold-GS optimizes only $(\vf^g)_{N'}$, the Gaussian offsets, and MLP weights, which significantly reduces parameters. This anchor-based formulation naturally yields a coarse partition of the Gaussian space, providing a strong basis for our subsequent clustering into \acrshort{superg}s.

\subsection{Segmentation Feature Field Distillation}
\label{sec:feats}
Given $N'$ anchor points representing the scene geometry, the next step is to group them into $S$ superpoints, each forming a \acrshort{superg} through its derived neural Gaussians. Ideally, each \acrshort{superg} should align with a single semantic entity in the scene. However, clustering anchors solely by their geometry features $\vf^g$ or positions $\vx$ is suboptimal, since anchors from distinct objects can be spatially adjacent or geometrically similar. To overcome this limitation, we introduce an additional segmentation feature $\vf^s$, distilled from 2D SAM masks, which encodes both instance- and part-level semantic cues to guide the \acrshort{superg} clustering.

\textbf{Hierarchical Partitioning of SAM Masks.} Given an input RGB image, SAM~\cite{kirillov2023sam} generates a set of 2D segmentation masks. These masks can, however, overlap with each other, leading to pixels belonging to multiple masks and thus obscuring the inherent part-instance hierarchy. Prior works either train separate models for each mask level~\cite{qin2024langsplat, cheng2024occam, goi2024}, which is less efficient, or rely only on coarse instance-level masks~\cite{wu2024opengaussian, drsplat25}, discarding the finer part-instance relations. To overcome this, we adopt a hierarchical representation~\cite{ying2024omniseg3d} that restructures the masks into non-overlapping instance-level masks $\mathcal{M}$ for whole objects and part-level patches $\mathcal{P}$ for finer components, which together provide supervision for learning both object-level semantics and intra-object details in the segmentation feature field. Implementation details and example mask visualizations are provided in Appendix~\ref{sec:more_impl}.

\begin{figure}[htpb]
    \centering 
    \includegraphics[width=0.9\linewidth]{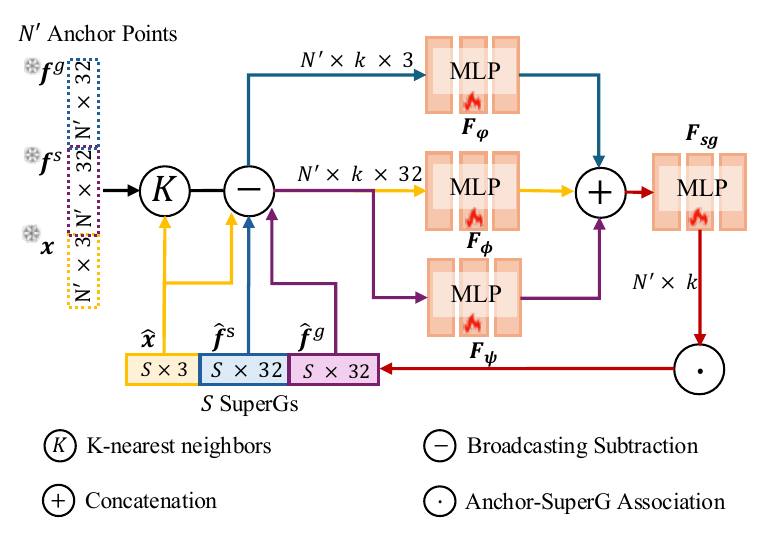} 
    \vspace{-0.3cm}
    \caption{The architecture of the \acrshort{superg} Cluster Network.}
    \label{fig:sg_module}
    \vspace{-0.5cm}
\end{figure}

\textbf{Instance and Hierarchical Feature Field.} 
As shown in~\figureautorefname~\ref{fig:overview}, we assign each anchor point a segmentation feature $\vf^s$. We pass $\vf^s$ together with the anchor position $\vx$ to a segmentation decoder to get the instance feature $\vg$ and hierarchical feature $\vh$ for each neural Gaussian. Through the vanilla Gaussian Splatting pipeline, we rasterize $\vg$ and $\vh$ to generate the 2D instance feature map $\hat{\mG} \in \sR^{D_g \times H \times W}$ and the 2D hierarchical feature map $\hat{\mH} \in \sR^{D_h \times H \times W}$.

To train the segmentation features, we leverage a contrastive learning objective~\cite{ying2024omniseg3d, silva2024contrastive} to enforce cross-view consistency, encouraging features from the same mask to be similar while pushing apart those from different masks.
Specifically, we represent the set of SAM-generated instance-level masks as $\mathcal{M}=\{{\vm}^p \in \sR^{H \times W} \mid p=1, \dots, |\mathcal{M}|\}$. Given an instance mask ${\vm}^p$, we collect all rendered instance features whose pixels fall inside the mask, and denote this set as $\hat{\vg}^p=\{\hat{\vg}^p_t \in \hat{\mG} \mid t = 1, \cdots, |\hat{\vg}^p|\}$. We compute the mean instance feature value within $\vm^p$ as $\bar{{\vg}}^p$ and the contrastive instance feature loss $\mathcal{L}_{Ins}$ is:

\begin{equation}
    \mathcal{L}_{Ins} = -\frac{1}{|\mathcal{M}|}\sum_{p=1}^{|\mathcal{M}|} \sum_{t=1}^{|\hat{\vg}^p|} 
    \log \frac{\exp \left(\hat{\vg}^p_t \cdot \bar{\vg}^p / \tau_p\right)}
    {\sum_{q=1}^{|\mathcal{M}|} 
    \exp\left(\hat{\vg}^p_t \cdot \bar{\vg}^q / \tau_q\right)},
\end{equation}
where $\tau$ is the cluster temperature. We adopt a similar hierarchical feature loss $\mathcal{L}_{Hier}$ from Omniseg3D~\cite{ying2024omniseg3d}, but applied to part-level patches $\mathcal{P}$ to supervise our hierarchical feature $\vh$. We refer to Appendix~\ref{sec:more_impl} for more details. Combined with the reconstruction loss introduced in \sectionautorefname~\ref{sec:pre}, these objectives define the overall training loss for Stage 1:

\begin{equation}
    \mathcal{L}_{stage1} = \mathcal{L}_{RGB} + \lambda_{Ins} \mathcal{L}_{Ins} + \lambda_{Hier} \mathcal{L}_{Hier}.
\end{equation}

\subsection{\acrlong{superg} Clustering}
\label{sec:Super-Gaussian}

After learning anchor-level geometry and segmentation features, we proceed to Stage 2, where anchors are grouped into semantically meaningful \acrshort{superg}s to form a compact representation. However, contrastive learning struggles to separate objects that never co-occur in training~\cite{kim2024garfield}, potentially grouping too distant Gaussians. To ensure spatial compactness and semantic consistency, we incorporate the anchor positions $\vx$ alongside segmentation features $\vf^s$, while geometric features $\vf^g$ provide appearance cues for refinement. A straightforward baseline is to apply $K$-means clustering~\cite{li2024instancegaussian} to the concatenated feature space of ${\{\vx, \vf^g, \vf^s\}}$. Yet, this approach fails when appearance cues misalign with semantics (e.g., diverse textures within an object). Moreover, $K$-means assumes equal importance across concatenated features, without the flexibility to adapt their relative relevance during clustering.To improve the clustering quality, we instead propose a learnable \acrshort{superg} clustering network~(see \figureautorefname~\ref{fig:sg_module}), inspired by~\cite{hui2021spnet}. It follows two steps: initialization and iterative refinement.

\textbf{\acrlong{superg} Initialization.} 
We apply the Farthest Point Sampling algorithm~\cite{qi2017pointnet++} on anchor points to initialize \acrshort{superg}s, averaging each a position $\widehat{\vx}$.
Each \acrshort{superg} has a geometry feature $\widehat{\vf}^g$ and segmentation feature $\widehat{\vf}^s$, which are initialized as the mean value of the corresponding anchors' features $\{\vf^g, \vf^s\}$.

\textbf{\acrlong{superg} Update.}
We denote the nearest $k$ \acrshort{superg}s to the $i$-th anchor as $\mathcal{N}_i$. The association probability matrix ${\mA} \in \mathbb{R}^{N' \times k}$~\cite{hui2021spnet, SP-GS} is used to weight the contribution of each \acrshort{superg} to its corresponding anchor, where $N'$ is the number of anchors and $k$ is the number of nearest \acrshort{superg}s. 
Specifically, the association probability between the $j$-th \acrshort{superg} ($j \in \mathcal{N}_i$) and the $i$-th anchor is:

\begin{equation}
    \mA_{ij} = F_{sg} \left( F_\phi(\vx_i, \widehat{\vx}_j) , F_\varphi(\vf^s_i, \widehat{\vf}^s_j) , F_\psi(\vf^g_i, \widehat{\vf}^g_j) \right),
\end{equation}
where $F_\phi$, $F_\varphi$, and $F_\psi$ are lightweight MLP decoders that output relevance weights in terms of spatial, semantic, and geometric information, respectively. The concatenated weights are then passed to the prediction decoder $F_{sg}$ for the normalized association probability matrix prediction.
Unlike $K$-means, this design dynamically adjusts the contribution of each \acrshort{superg} to its corresponding anchor.

We iteratively update \acrshort{superg}s through the association matrix ${\mA}$. At iteration $t+1$, each \acrshort{superg}'s position and features are updated with its corresponding anchors:
\begin{equation}
    \widehat{\mathbf{e}}_j^{t+1} = \frac{1}{\sum_{i=1}^{N'} \mathbb{I}(j \in \mathcal{N}_i) {\mA}_{ij}^t } \sum_{i=1}^{N'} \mathbb{I}(j \in \mathcal{N}_i) {\mA}_{ij}^t \mathbf{e}_i,
\end{equation}
where $\mathbb{I}$ denotes the indicator function, $\mathbf{e} \in {\{\vx, \vf^g, \vf^s\}}$ are the anchor's attributes and $\widehat{\mathbf{e}} \in {\{\widehat{\vx}, \widehat{\vf}^g, \widehat{\vf}^s\}}$ are \acrshort{superg}'s.

We optimize the \acrshort{superg} clustering network to learn the association matrix ${\mA}$, ensuring that the derived \acrshort{superg} attributes $\widehat{\mathbf{e}}$ accurately reconstruct the anchor attributes $\mathbf{e}$. Note that $\mathbf{e}$ from Stage 1~(Section~\ref{sec:feats}) are now frozen:
\begin{equation}
    \mathcal{L}_{recon, \mathbf{e}} = \frac{1}{N'} \sum_{i=1}^{N'} \lVert \mathbf{e}_i - \sum_{j \in \mathcal{N}_i} \mA_{ij} \widehat{\mathbf{e}}_j \rVert.
\end{equation}

However, anchors within the same \acrshort{superg} may be semantically similar yet spatially distant, especially when contrastive learning fails to optimize instances that never co-occur in the same view. To enforce spatial coherence, we introduce a compactness objective:

\begin{equation}
    \mathcal{L}_{compact, \mathcal{X}} = \frac{1}{S} \sum_{j=1}^S \sum_{\mathbf{x} \in \mathcal{X}_j} \lVert \mathbf{x} - \widehat{\vx}_j \rVert,
\end{equation}
where $\mathcal{X}_j$ is the set of anchors' position assigned to the $j$-th \acrshort{superg}. This loss encourages assigned anchors to cluster around their \acrshort{superg} center and avoid fragmentation.

\begin{figure*}[t]
    \centering 
    \includegraphics[width=15cm]{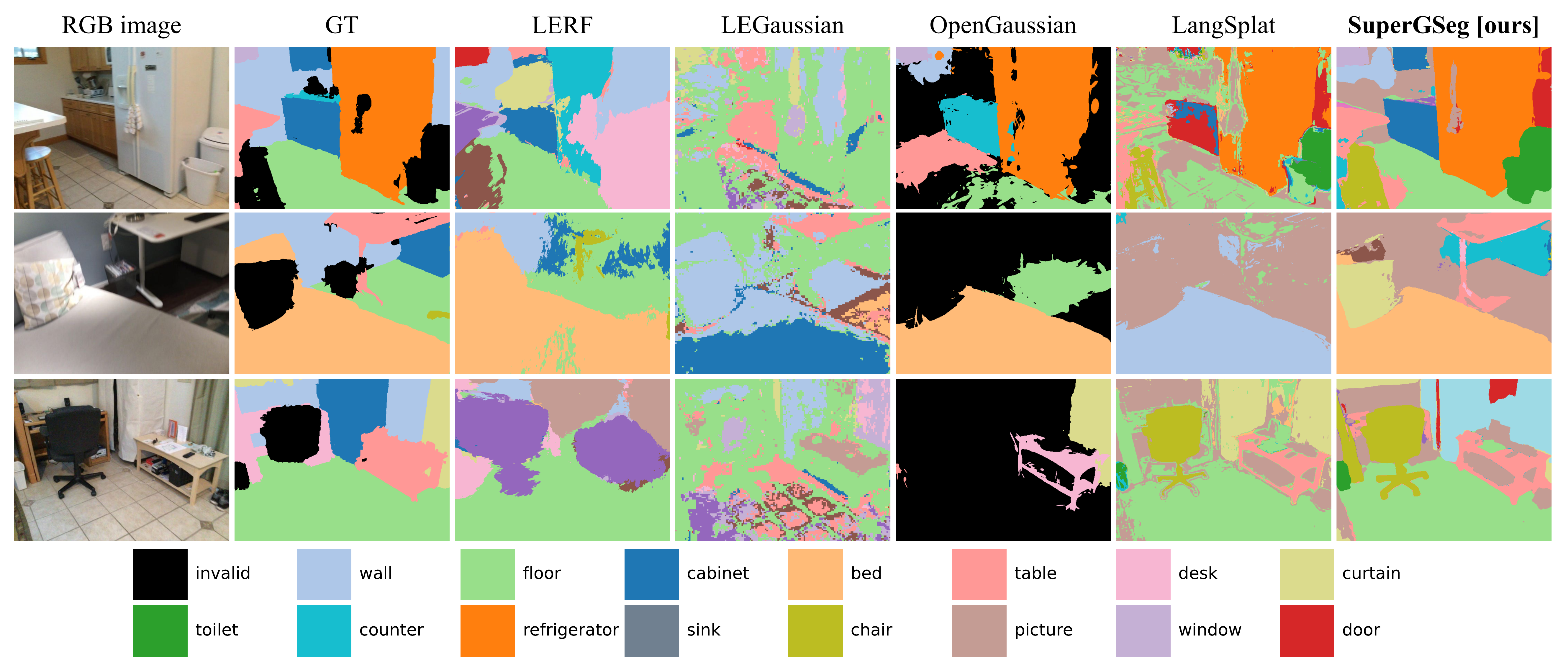} 
    \vspace{-0.2cm}
    \caption{Qualitative comparison of semantic segmentation predictions on the ScanNet v2 dataset~\cite{dai2017scannet}. }
    \label{fig:scannet_qualitative}
\end{figure*}

\begin{table*}[ht]
\setlength{\tabcolsep}{2.68pt}
\centering
\begin{tabular}{l|cc|cc|cc|cc|cc|cc|cc}
\toprule
 & \multicolumn{2}{c|}{\textbf{mean}} & \multicolumn{2}{c|}{\textit{wall}} & \multicolumn{2}{c|}{\textit{floor}} & \multicolumn{2}{c|}{\textit{cabinet}} & \multicolumn{2}{c|}{\textit{chair}} & \multicolumn{2}{c|}{\textit{refrigerator}} & \multicolumn{2}{c}{\textit{curtain}} \\
Method & mIoU & mAcc & mIoU & mAcc & mIoU & mAcc & mIoU & mAcc & mIoU & mAcc & mIoU & mAcc & mIoU & mAcc \\
\midrule
LERF \cite{kerr2023lerf} &\cellcolor{orange!20}38.5 & \cellcolor{yellow!20}60.4& 35.2& 82.8& 60.1& 68.8& 52.0&82.7& 10.9& 10.9&   69.9 & 90.2& 70.2& 77.8\\
LEGaussians \cite{shi2024language}    & 8.7  & 33.2 & 17.9 & 53.1 & 14.6 & 20.6 & 2.7  & 18.6 & 0.4  & 28.7 & 9.0  & 74.3  & 1.9  & 10.4 \\
OpenGaussian \cite{wu2024opengaussian}& \cellcolor{yellow!20}24.1 & \cellcolor{orange!20}68.7 & 13.4 & \textbf{96.6} & 31.2 & 74.4 & 0.3  & 22.9 & 36.5 & 83.4  &  \textbf{88.0} & \textbf{98.3} & 17.7 & \textbf{79.2} \\
LangSplat \cite{qin2024langsplat}      & 27.6 & 48.3 & 45.3 & 72.6 & 43.3 & 45.6 & 24.8 & 56.7 & 18.0 & 48.5 &  0.7  & 33.3 & 46.8 & 66.5 \\
\textbf{SuperGSeg} [ours]      & \cellcolor{red!20}\textbf{54.7} & \cellcolor{red!20}\textbf{74.7} & \textbf{58.8} & 92.9 & \textbf{53.6} & \textbf{86.5} & \textbf{69.8} & \textbf{83.8} & \textbf{80.4} & \textbf{83.8} & 79.4 & 80.2& \textbf{61.8} & 64.5 \\
\bottomrule
\end{tabular}
\vspace{-2mm}
\caption{Comparison on the ScanNet v2 dataset~\cite{dai2017scannet}. We report the mean result and detailed scores for the most common object categories, following the evaluation protocol of~\cite{chen2024panoptic}. Results for more categories are provided in the Appendix~\ref{sec:more_result}.}
\label{tab:scannet}
\vspace{-2mm}
\end{table*}

\subsection{Language Field Distillation}
\label{sec:lang_field}

Building on the clustering from Stage 2 (\sectionautorefname~\ref{sec:Super-Gaussian}), in Stage 3, we distill 2D CLIP features into our compact set of $S$ \acrshort{superg}s, rather than into millions of individual 3D Gaussians to enable open-vocabulary 3D scene understanding. This design ensures consistent, robust, and high-dimensional language representations, while avoiding the feature degradation typically caused by the lossy compression used in Gaussian-based distillation approaches.

Since all Gaussians within a \acrshort{superg} are expected to share the same semantics, we assign each \acrshort{superg} a learnable latent language feature $\widehat{\vf}^l$. As shown in~\figureautorefname~\ref{fig:overview}, this latent feature, together with the \acrshort{superg} position $\widehat{\vx}$, is decoded by a language feature MLP $F_L$ to produce a CLIP-aligned feature: $\widehat{\vl} = F_L(\widehat{\vf}^l, \widehat{\vx})$. We then modify the rasterizer to render a language feature map $\hat{\mL}$, using $\widehat{\vl}$ and the anchor-\acrshort{superg} association map $\mA$. For supervision, instance masks obtained in~\sectionautorefname~\ref{sec:feats} are encoded using the CLIP image encoder to produce target 2D CLIP features $\mL$. The latent features $\widehat{\vf}^l$ and the decoder $F_L$ are jointly optimized using a cosine similarity loss:
\begin{equation}
    \mathcal{L}_{Lang} = 1 - \cos(\hat{\mL}, \mL).
\end{equation}

\section{Experiments}
\label{sec:experiment}

\begin{figure*}[t]
    \centering 
    \includegraphics[width=15cm]{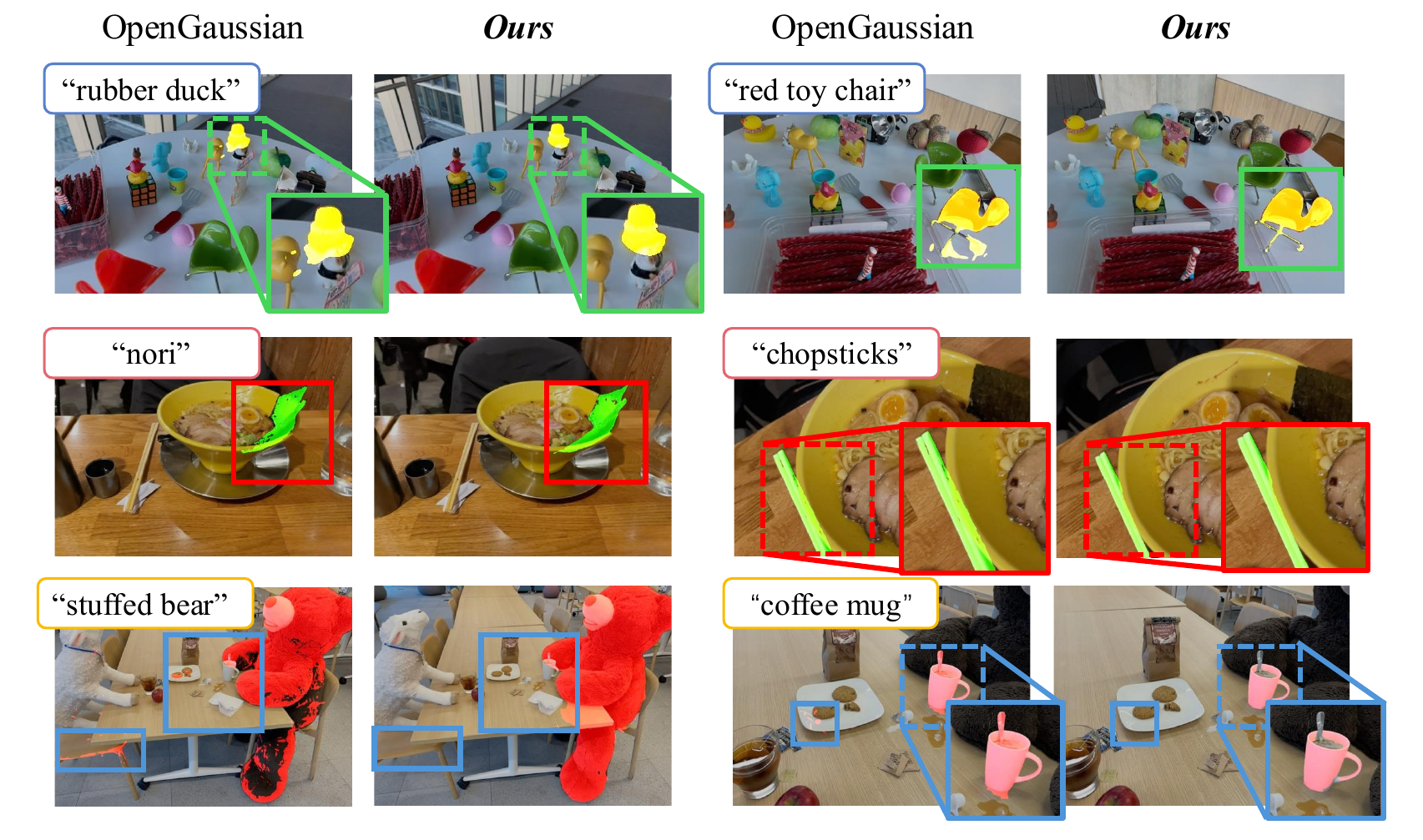} 
    \vspace{-0.5cm}
    \caption{Qualitative comparison on the LERF-OVS dataset~\cite{kerr2023lerf} for the open-vocabulary 3D object selection task. Text queries for each scene are displayed in quotation marks. SuperGSeg delivers more precise and less noisy segmentation masks.}
    \label{fig:lerf_qualitative}
\end{figure*}

\begin{table*}[ht]
\setlength{\tabcolsep}{4.5pt}
\centering
\begin{tabular}{l|cc|cc|cc|cc|cc|cc}
\toprule
&\multicolumn{2}{c|}{Inference}  & \multicolumn{2}{c|}{\textbf{mean}} & \multicolumn{2}{c|}{\textit{figurines}} & \multicolumn{2}{c|}{\textit{teatime}} & \multicolumn{2}{c|}{\textit{ramen}} & \multicolumn{2}{c}{\textit{waldo\_kitchen}} \\
Method & Time& Mem. &mIoU & mAcc & mIoU & mAcc & mIoU & mAcc & mIoU & mAcc & mIoU & mAcc \\
\midrule
LangSplat~\cite{qin2024langsplat} & \cellcolor{orange!20}3.28s & 18GB &9.66 & 12.41 & 10.16 & 8.93 & 11.38 & 20.34 & 7.92 & 11.27 & 9.18 & 9.09 \\
LEGaussians~\cite{shi2024language} & \cellcolor{yellow!20}4.42s & \cellcolor{orange!20}5GB & \cellcolor{yellow!20}16.21 & \cellcolor{yellow!20}23.82 & 17.99 & 23.21 & 19.27 & 27.12 & 15.79 & 26.76 & 11.78 & 18.18 \\
OpenGaussian~\cite{wu2024opengaussian} & 5.55s & \cellcolor{yellow!20}{9GB} & \cellcolor{red!20}\textbf{38.36} & \cellcolor{orange!20}51.43 & 39.29 & 55.36 & \textbf{60.44} & 76.27 & \textbf{31.01} & \textbf{42.25} & 22.70 & 31.82 \\
\textbf{SuperGSeg} [ours] & \cellcolor{red!20}\textbf{0.50s} & \cellcolor{red!20}\textbf{4GB} & \cellcolor{orange!20}35.94 & \cellcolor{red!20}\textbf{52.02} & \textbf{43.68} & \textbf{60.71} & 55.31 & \textbf{77.97} & 18.07 & 23.94 & \textbf{26.71} & \textbf{45.45} \\
\bottomrule
\end{tabular}
\vspace{-2mm}
\caption{Open-vocabulary 3D object selection comparison on the LERF-OVS dataset~\cite{qin2024langsplat}. LERF~\cite{kerr2023lerf} is not applicable for this task. We report the \acrshort{mIoU} and \acrshort{mAcc} of compared methods as provided in~\cite{wu2024opengaussian}, and measure inference cost using their official implementations.}
\label{tab:lerf}
\end{table*}

\subsection{Experimental Setup}

\textbf{Datasets.} We evaluate our method on the \textbf{open-vocabulary novel view semantic segmentation and object selection tasks} using the ScanNet v2~\cite{dai2017scannet} and LERF-OVS~\cite{qin2024langsplat} datasets. ScanNet v2~\cite{dai2017scannet} includes posed RGB images and 2D semantic labels of indoor scenes. We randomly select 8 scenes from the dataset. These include a variety of indoor environments, e.g., living rooms, bedrooms, kitchens, and offices. For each scene, we split the data into a training set (composed of every 20th image from the original sequence) and a test set (derived from the intermediate images between the training set samples). For semantic segmentation, we specifically use the 20 object categories. LERF-OVS~\cite{qin2024langsplat} consists of complex in-the-wild scenes captured with consumer-level devices, annotated with ground truth masks of textual queries to enable evaluation for open-vocabulary object selection tasks.

\textbf{Baselines and Metrics.} We compare our method with representative NeRF-based and 3DGS-based baselines, including  LERF~\cite{kerr2023lerf}, LangSplat~\cite{qin2024langsplat}, LEGaussian~\cite{shi2024language},  and OpenGaussian~\cite{wu2024opengaussian}. For the open-vocabulary semantic segmentation task, CLIP-encoded text features are compared with rendered 2D language feature maps via cosine similarity to produce per-pixel semantic predictions~\cite{zhou2024feature}, evaluated with \acrlong{mIoU} (\acrshort{mIoU}) and \acrlong{mAcc} (\acrshort{mAcc}). For the open-vocabulary object selection task, we perform text queries directly in 3D space~\cite{wu2024opengaussian}, retrieving the most relevant \acrshort{superg}s and rendering them into 2D for evaluation with \acrshort{mIoU} and \acrshort{mAcc}. Since NeRF is an implicit representation without explicit 3D positions, LERF cannot be applied to this task. We also report inference-time efficiency, measuring both runtime and memory consumption for text queries on trained 3D scenes. Specifically, we perform multiple queries from different viewpoints and report the average query time. We consider this metric particularly important for assessing the feasibility of deploying models on resource-constrained devices and enabling real-time querying in practical scenarios.

\textbf{Implementation Details.}
The training process is divided into 3 stages. In the first stage, we train the Scaffold-GS~\cite{lu2024scaffold} with instance and hierarchical features for 30k iterations. In the second stage, we freeze the geometry and segmentation features from stage one and train only the \acrshort{superg} clustering network for another 30k iterations. In the last stage, we freeze all other parameters and optimize the language features for each \acrshort{superg} for 10k iterations. For more implementation details, we refer to Appendix~\ref{sec:more_impl}.

\subsection{Open-Vocabulary Semantic Segmentation}

\label{sec:scannet}
\textbf{Quantitative Results.} 
As shown in \tableautorefname~\ref{tab:scannet}, SuperGSeg achieves the best overall scores in both \acrshort{mIoU} and \acrshort{mAcc} among the compared methods, demonstrating its effectiveness in capturing the open-set information of the scene, yielding remarkable performance in a variety of object categories. In comparison, LEGaussian~\cite{li2024geogaussian} shows lower performance on both metrics, suggesting limited generalization across multiple object categories. LangSplat~\cite{qin2024langsplat} performs better than LEGaussian but still shows reduced accuracy in more diverse categories. OpenGaussian~\cite{wu2024opengaussian} obtains competitive results on certain large structures such as wall and floor, but its overall scene-level performance remains below ours. LERF~\cite{kerr2023lerf} achieves the second-highest mIoU, though its relatively low \acrshort{mAcc} suggests difficulties in producing clear segmentation boundaries.

\textbf{Qualitative Results.} 
As shown in \figureautorefname~\ref{fig:scannet_qualitative}, our method produces sharper and more semantically consistent masks than the compared methods. While OpenGaussian~\cite{wu2024opengaussian} demonstrates competitive performance in 3D object-level semantic segmentation (\sectionautorefname~\ref{sec:lerf_ovs}), it struggles in dense pixel-wise semantic segmentation. This is evident with occlusions due to projections onto 2D-pixel space. Without alpha blending, the occluded Gaussians cannot be effectively distinguished from one another. Instead, LangSplat~\cite{qin2024langsplat} produces fine border segmentation but often includes incorrect semantic labels and noisy predictions, likely due to the lossy encoding of language information. LERF~\cite{kerr2023lerf} presents accurate semantic prediction but with imprecise boundaries, limiting its applicability in fine-grained segmentation tasks.

\subsection{Open-Vocabulary Object Selection}

\label{sec:lerf_ovs}
\textbf{Quantitative Results.} 
SuperGSeg improves over baseline methods that assign and optimize language features per Gaussian~\cite{zhou2024feature, qin2024langsplat, li2024geogaussian}. As shown in \tableautorefname~\ref{tab:lerf}, clustering Gaussians into \acrshort{superg}s enhances both spatial and semantic accuracy over per-Gaussian methods. We further compare SuperGSeg to OpenGaussian~\cite{wu2024opengaussian}, another method exploring 3D Gaussian clustering. OpenGaussian’s direct 2D CLIP feature association yields a slightly higher mIoU by avoiding alpha-blending artifacts, but it underperforms in 2D semantic segmentation on ScanNet (\sectionautorefname~\ref{sec:scannet}). In contrast, SuperGSeg maintains competitive mIoU for 3D object selection while surpassing OpenGaussian in 2D semantic segmentation, enhancing its versatility across real-world applications. Our higher mAcc, especially in complex LERF-OVS scenes such as \textit{figurines} and \textit{waldo kitchen}, reflects the precision of \acrlong{superg} clustering and instance grouping. By accurately segmenting Gaussians in 3D, SuperGSeg renders more complete 2D masks with sharper boundaries, improving semantic consistency in challenging settings.
In addition, SuperGSeg reduces inference latency to around 0.5s per query and decreases memory usage to 4GB, more than 50\% lower than the next best baseline at 9GB. These improvements, enabled by \acrshort{superg}s, demonstrate the potential for real-time querying on resource-constrained devices.

\textbf{Qualitative Results.} For visualization, we query language features in 3D space and render the resulting 3D masks to 2D. As shown in~\figureautorefname~\ref{fig:lerf_qualitative}, SuperGSeg delivers precise 3D object selection without spurious outliers and produces clearer boundaries. Thanks to the 3D understanding capability, our SuperGSeg allows for effective localization of occluded regions (e.g., the \textit{stuffed bear} leg under a table). Notably, its high-quality features distinguish the coffee mug from its contents and spoon, showcasing the efficacy of distilling fine-grained features into \acrshort{superg}s.

\textbf{Ablation Study.} We conduct ablation studies on various components of our method to validate the necessity of \acrshort{superg}s, as summarized in \tableautorefname~\ref{table:ablation_sg}. 
The baseline without \acrshort{superg} (case a) trains the language feature field by directly optimizing per-anchor features, which results in limited semantic consistency. 
To analyze how different feature types affect \acrshort{superg} formation, we evaluate grouping based solely on anchor coordinates and geometric features (case b), instance features (case c), and hierarchical features (case d). 
The results indicate that grouping Gaussians into \acrshort{superg} improves semantic consistency compared to per-anchor optimization, but relying only on coordinates and geometry remains suboptimal. 
Both instance and hierarchical features contribute substantially to accurate \acrshort{superg} assignments, and the best performance is achieved with our full model (case f), which combines both. 
We further compare $K$-means clustering for Gaussian grouping (case e) with our learnable \acrshort{superg} assignment (case f). 
By dynamically adapting to variations in the feature space, our learnable predictor produces higher-quality \acrshort{superg}s, yielding consistently higher \acrshort{mIoU} and improved \acrshort{mAcc}. Additional ablation studies on components of the \acrshort{superg} clustering network are provided in Appendix~\ref{sec:more_abl}.

\begin{table}[htpb]
\setlength{\tabcolsep}{3pt}
\vspace{-2mm}
  \resizebox{\linewidth}{!}{
  \centering
  \begin{tabular}{cccc|cc}
    \toprule
    \# & w/ Learned \acrshort{superg} & w/ ins & w/ hier & mIoU $\uparrow$ & mAcc. $\uparrow$ \\
    \midrule
    a)& \ & \ & \ & 10.12 & 14.49 \\
    b) & \checkmark & \ & \ & 12.08 & 16.95 \\
    c) & \checkmark & \checkmark & \ & 53.91 & 64.41 \\
    d) & \checkmark & \ & \checkmark & 49.04 & 66.10 \\
    e) & \ & \checkmark & \checkmark & 53.77 & 67.80\\
    f) & \checkmark & \checkmark & \checkmark & \textbf{55.31} & \textbf{77.97} \\
    \bottomrule
  \end{tabular}
  }
  \caption{\acrshort{superg} ablation study, \textit{teatime} scene of LERF-OVS.}
  \label{table:ablation_sg}
\end{table}

\subsection{Application}

\begin{figure}[htpb]
    \vspace{-0.2cm}
    \centering 
    \includegraphics[width=0.9\linewidth]{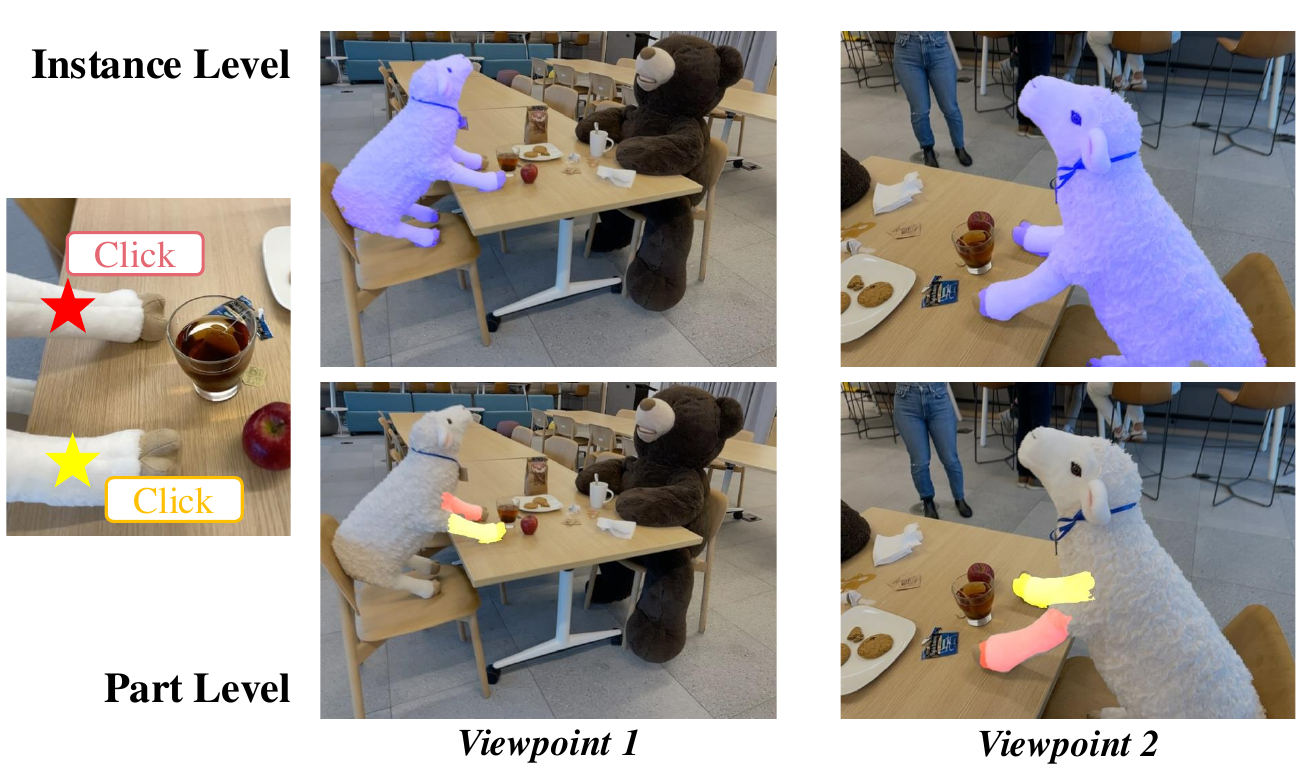} 
    \vspace{-0.3cm}
    \caption{Cross-level and cross-frame segmentation visualization.}
    \label{fig:cross_level_frame_query}
    \vspace{-0.8cm}
\end{figure}

Beyond language-based querying, \acrshort{superg}s serve as a multi-granularity representation of 3D scenes by integrating instance- and part-level knowledge, readily applicable to tasks such as cross-frame segmentation and hierarchical instance decomposition, without requiring task-specific retraining. For example, a click on a reference image retrieves \acrshort{superg}s with matching hierarchical features, allowing the selected part to be consistently rendered across views. In addition to cross-view querying, SuperGSeg enables cross-level queries: clicking on a part retrieves its parent object using instance features, while clicking on an object reveals its constituent parts, which supports seamless navigation from parts to instances and vice versa, as illustrated in~\figureautorefname~\ref{fig:cross_level_frame_query}. Furthermore, the granularity of instance-to-part segmentation can be adjusted by varying the threshold on hierarchical feature similarity, as shown in~\figureautorefname~\ref{fig:intra_hierarchy}. Additional implementation details are provided in Appendix~\ref{sec:superg_details}.

\begin{figure}[htpb]
    \centering 
    \includegraphics[width=0.9\linewidth]{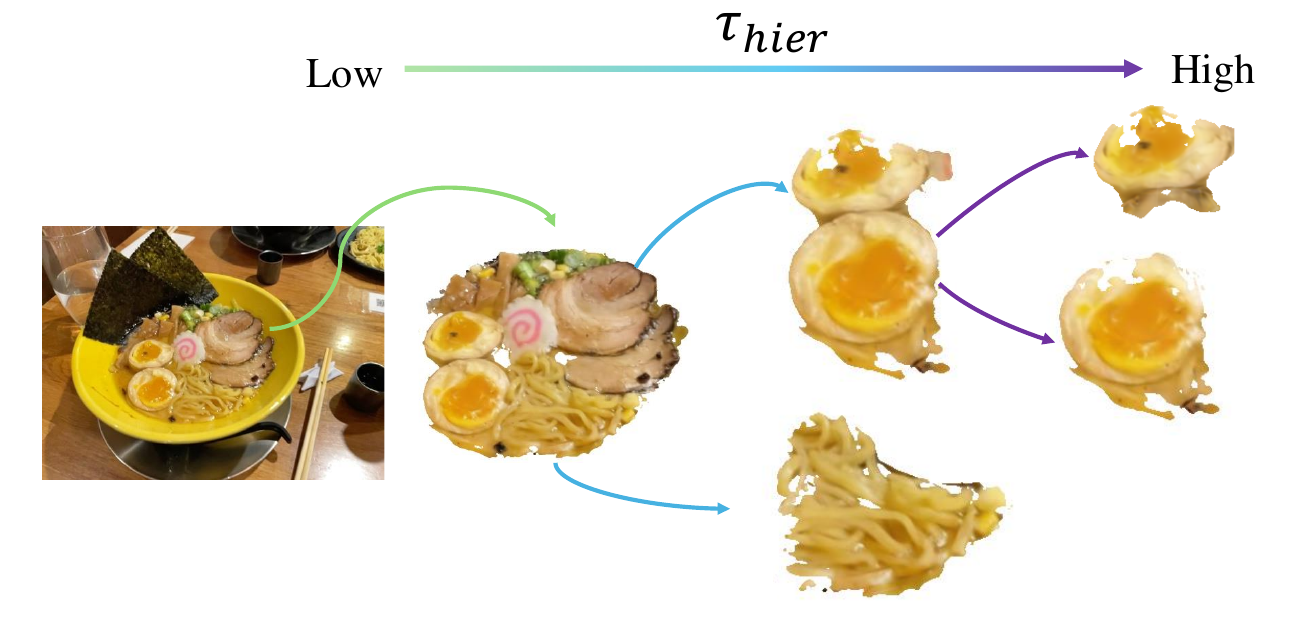} 
    \vspace{-0.5cm}
    \caption{Visualization of intra-object hierarchy definition.}
    \label{fig:intra_hierarchy}
    \vspace{-0.6cm}
\end{figure}

\section{Conclusion}
\label{sec:conclusion}

We present SuperGSeg, a novel framework for 3D scene understanding that represents scenes using compact \acrlong{superg}s, ensuring semantic and appearance consistency. By leveraging neural Gaussians, our method captures instance- and part-level segmentation features, guiding \acrlong{superg} clustering through an adaptive online learning algorithm. Experiments show that integrating high-dimensional language features significantly improves open-set 3D language querying, demonstrating the framework's remarkable performance. Furthermore, the \acrlong{superg} representation is readily adaptable to a wide range of 3D scene understanding tasks.

\section{Acknowledgement}
\label{sec:acknowledgement}
This work was supported by the German Research Foundation (DFG): SFB 1233, Robust Vision: Inference Principles and Neural Mechanisms, TP A1, project number: 276693517, and the Machine Learning Cluster of Excellence, EXC number 2064/1 – Project number 390727645. The authors thank the International Max Planck Research School for Intelligent Systems (IMPRS-IS) for supporting Siyun Liang.

{
    \small
    \bibliographystyle{unsrt}
    \bibliography{main}

@String(CVPR= {IEEE Conf. Comput. Vis. Pattern Recog.})

@String(ICCV= {Int. Conf. Comput. Vis.})

@String(ICLR = {Int. Conf. Learn. Represent.})

@String(AAAI = {AAAI})

@String(CVPR  = {CVPR})

@String(ICCV  = {ICCV})

@String(ICLR  = {ICLR})

@article{kerbl20233d,
  title={{3D Gaussian Splatting} for Real-Time Radiance Field Rendering.},
  author={Kerbl, Bernhard and Kopanas, Georgios and Leimk{\"u}hler, Thomas and Drettakis, George},
  journal={ACM Transactions on Graphics},
  volume={42},
  number={4},
  pages={139--1},
  year={2023}
}

@inproceedings{li2024geogaussian,
  title={Geogaussian: Geometry-aware gaussian splatting for scene rendering},
  author={Li, Yanyan and Lyu, Chenyu and Di, Yan and Zhai, Guangyao and Lee, Gim Hee and Tombari, Federico},
  booktitle={European Conference on Computer Vision},
  pages={441--457},
  year={2024},
  organization={Springer}
}

@article{yu2024gaussian,
  title={Gaussian opacity fields: Efficient and compact surface reconstruction in unbounded scenes},
  author={Yu, Zehao and Sattler, Torsten and Geiger, Andreas},
  journal={arXiv preprint arXiv:2404.10772},
  year={2024}
}

@inproceedings{dai2024high,
  title={High-quality surface reconstruction using gaussian surfels},
  author={Dai, Pinxuan and Xu, Jiamin and Xie, Wenxiang and Liu, Xinguo and Wang, Huamin and Xu, Weiwei},
  booktitle={ACM SIGGRAPH 2024 Conference Papers},
  pages={1--11},
  year={2024}
}

@inproceedings{guedon2024sugar,
  title={Sugar: Surface-aligned gaussian splatting for efficient 3d mesh reconstruction and high-quality mesh rendering},
  author={Gu{\'e}don, Antoine and Lepetit, Vincent},
  booktitle={Proceedings of the IEEE/CVF Conference on Computer Vision and Pattern Recognition},
  pages={5354--5363},
  year={2024}
}

@inproceedings{lu2024scaffold,
  title={Scaffold-gs: Structured 3d gaussians for view-adaptive rendering},
  author={Lu, Tao and Yu, Mulin and Xu, Linning and Xiangli, Yuanbo and Wang, Limin and Lin, Dahua and Dai, Bo},
  booktitle={Proceedings of the IEEE/CVF Conference on Computer Vision and Pattern Recognition},
  pages={20654--20664},
  year={2024}
}

@inproceedings{kerr2023lerf,
  title={Lerf: Language embedded radiance fields},
  author={Kerr, Justin and Kim, Chung Min and Goldberg, Ken and Kanazawa, Angjoo and Tancik, Matthew},
  booktitle={Proceedings of the IEEE/CVF International Conference on Computer Vision},
  pages={19729--19739},
  year={2023}
}

@inproceedings{radford2021clip,
  title={Learning transferable visual models from natural language supervision},
  author={Radford, Alec and Kim, Jong Wook and Hallacy, Chris and Ramesh, Aditya and Goh, Gabriel and Agarwal, Sandhini and Sastry, Girish and Askell, Amanda and Mishkin, Pamela and Clark, Jack and others},
  booktitle={Proceedings of the International Conference on Machine Learning},
  pages={8748--8763},
  year={2021},
  organization={PMLR}
}

@inproceedings{kirillov2023sam,
  title={Segment anything},
  author={Kirillov, Alexander and Mintun, Eric and Ravi, Nikhila and Mao, Hanzi and Rolland, Chloe and Gustafson, Laura and Xiao, Tete and Whitehead, Spencer and Berg, Alexander C and Lo, Wan-Yen and others},
  booktitle={Proceedings of the IEEE/CVF International Conference on Computer Vision},
  pages={4015--4026},
  year={2023}
}

@inproceedings{ying2024omniseg3d,
  title={Omniseg{3D}: Omniversal {3D} segmentation via hierarchical contrastive learning},
  author={Ying, Haiyang and Yin, Yixuan and Zhang, Jinzhi and Wang, Fan and Yu, Tao and Huang, Ruqi and Fang, Lu},
  booktitle={Proceedings of the IEEE/CVF Conference on Computer Vision and Pattern Recognition},
  pages={20612--20622},
  year={2024}
}

@inproceedings{caron2021dino,
  title={Emerging Properties in Self-Supervised Vision Transformers},
  author={Caron, Mathilde and Touvron, Hugo and Misra, Ishan and J\'egou, Herv\'e  and Mairal, Julien and Bojanowski, Piotr and Joulin, Armand},
  booktitle={Proceedings of the IEEE/CVF International Conference on Computer Vision},
  year={2021}
}

@inproceedings{qin2024langsplat,
  title={Langsplat: {3D} language gaussian splatting},
  author={Qin, Minghan and Li, Wanhua and Zhou, Jiawei and Wang, Haoqian and Pfister, Hanspeter},
  booktitle={Proceedings of the IEEE/CVF Conference on Computer Vision and Pattern Recognition},
  pages={20051--20060},
  year={2024}
}

@inproceedings{mildenhall2020nerf,
  title={{NeRF}: Representing Scenes as Neural Radiance Fields for View Synthesis},
  author={Mildenhall, Ben and Srinivasan, Pratul P and Tancik, Matthew and Barron, Jonathan T and Ramamoorthi, Ravi and Ng, Ren},
  booktitle={Proceedings of the European Conference on Computer Vision},
  pages={405--421},
  year={2020}
}

@inproceedings{zhou2024feature,
  title={Feature {3DGS}: Supercharging {3D} gaussian splatting to enable distilled feature fields},
  author={Zhou, Shijie and Chang, Haoran and Jiang, Sicheng and Fan, Zhiwen and Zhu, Zehao and Xu, Dejia and Chari, Pradyumna and You, Suya and Wang, Zhangyang and Kadambi, Achuta},
  booktitle={Proceedings of the IEEE/CVF Conference on Computer Vision and Pattern Recognition},
  pages={21676--21685},
  year={2024}
}

@inproceedings{gaussian_grouping,
    title={Gaussian Grouping: Segment and Edit Anything in {3D} Scenes},
    author={Ye, Mingqiao and Danelljan, Martin and Yu, Fisher and Ke, Lei},
    booktitle={Proceedings of the European Conference on Computer Vision},
    pages={162--179},
    year={2024}
}

@article{wu2024opengaussian,
    title={OpenGaussian: Towards Point-Level {3D} Gaussian-based Open Vocabulary Understanding},
    author={Wu, Yanmin and Meng, Jiarui and Li, Haijie and Wu, Chenming and Shi, Yahao and Cheng, Xinhua and Zhao, Chen and Feng, Haocheng and Ding, Errui and Wang, Jingdong and others},
    journal={arXiv preprint arXiv:2406.02058},
    year={2024}
}

@article{yi2024gaussiandreamerpro,
  title={GaussianDreamerPro: Text to Manipulable {3D} Gaussians with Highly Enhanced Quality},
  author={Yi, Taoran and Fang, Jiemin and Zhou, Zanwei and Wang, Junjie and Wu, Guanjun and Xie, Lingxi and Zhang, Xiaopeng and Liu, Wenyu and Wang, Xinggang and Tian, Qi},
  journal={arXiv preprint arXiv:2406.18462},
  year={2024}
}

@inproceedings{dai2017scannet,
    title={ScanNet: Richly-annotated {3D} Reconstructions of Indoor Scenes},
    author={Dai, Angela and Chang, Angel X. and Savva, Manolis and Halber, Maciej and Funkhouser, Thomas and Nie{\ss}ner, Matthias},
    booktitle = {Proceedings of the IEEE/CVF Conference on Computer Vision and Pattern Recognition},
    year = {2017}
}

@inproceedings{shi2024language,
  title={Language embedded {3D} gaussians for open-vocabulary scene understanding},
  author={Shi, Jin-Chuan and Wang, Miao and Duan, Hao-Bin and Guan, Shao-Hua},
  booktitle={Proceedings of the IEEE/CVF Conference on Computer Vision and Pattern Recognition},
  pages={5333--5343},
  year={2024}
}

@inproceedings{
li2022languagedriven,
title={Language-driven Semantic Segmentation},
author={Boyi Li and Kilian Q Weinberger and Serge Belongie and Vladlen Koltun and Rene Ranftl},
booktitle={International Conference on Learning Representations},
year={2022},
url={https://openreview.net/forum?id=RriDjddCLN}
}

@InProceedings{KingBa15,
  author    = {Kingma, Diederik and Ba, Jimmy},
  booktitle = {International Conference on Learning Representations (ICLR)},
  title     = {Adam: A Method for Stochastic Optimization},
  year      = {2015},
}

@inproceedings{fridovich2022plenoxels,
  title={Plenoxels: Radiance fields without neural networks},
  author={Fridovich-Keil, Sara and Yu, Alex and Tancik, Matthew and Chen, Qinhong and Recht, Benjamin and Kanazawa, Angjoo},
  booktitle={Proceedings of the IEEE/CVF conference on computer vision and pattern recognition},
  pages={5501--5510},
  year={2022}
}

@article{zuo2024fmgs,
  title={Fmgs: Foundation model embedded {3D} gaussian splatting for holistic {3D} scene understanding},
  author={Zuo, Xingxing and Samangouei, Pouya and Zhou, Yunwen and Di, Yan and Li, Mingyang},
  journal={International Journal of Computer Vision},
  pages={1--17},
  year={2024},
  publisher={Springer}
}

@article{zwicker2002ewa,
  title={EWA splatting},
  author={Zwicker, Matthias and Pfister, Hanspeter and Van Baar, Jeroen and Gross, Markus},
  journal={IEEE Transactions on Visualization and Computer Graphics},
  volume={8},
  number={3},
  pages={223--238},
  year={2002},
  publisher={IEEE}
}

@inproceedings{hui2021spnet,
  title={Superpoint Network for Point Cloud Oversegmentation},
  author={Hui, Le and Yuan, Jia and Cheng, Mingmei and Xie, Jin and Yang, Jian},
  booktitle={ICCV},
  year={2021}
}

@InProceedings{SP-GS,
  title = 	 {Superpoint Gaussian Splatting for Real-Time High-Fidelity Dynamic Scene Reconstruction},
  author =       {Wan, Diwen and Lu, Ruijie and Zeng, Gang},
  booktitle = 	 {Proceedings of the 41st International Conference on Machine Learning},
  pages = 	 {49957--49972},
  year = 	 {2024},
}

@inproceedings{landrieu2018large,
  title={Large-scale point cloud semantic segmentation with superpoint graphs},
  author={Landrieu, Loic and Simonovsky, Martin},
  booktitle={Proceedings of the IEEE conference on computer vision and pattern recognition},
  pages={4558--4567},
  year={2018}
}

@inproceedings{landrieu2019point,
  title={Point cloud oversegmentation with graph-structured deep metric learning},
  author={Landrieu, Loic and Boussaha, Mohamed},
  booktitle={Proceedings of the IEEE/CVF Conference on Computer Vision and Pattern Recognition},
  pages={7440--7449},
  year={2019}
}

@inproceedings{robert2023efficient,
  title={Efficient {3D} semantic segmentation with superpoint transformer},
  author={Robert, Damien and Raguet, Hugo and Landrieu, Loic},
  booktitle={Proceedings of the IEEE/CVF International Conference on Computer Vision},
  pages={17195--17204},
  year={2023}
}

@inproceedings{robert2024scalable,
  title={Scalable {3D} Panoptic Segmentation As Superpoint Graph Clustering},
  author={Robert, Damien and Raguet, Hugo and Landrieu, Loic},
  booktitle={2024 International Conference on 3D Vision (3DV)},
  pages={179--189},
  year={2024},
  organization={IEEE}
}

@inproceedings{papon2013voxel,
  title={Voxel cloud connectivity segmentation-supervoxels for point clouds},
  author={Papon, Jeremie and Abramov, Alexey and Schoeler, Markus and Worgotter, Florentin},
  booktitle={Proceedings of the IEEE conference on computer vision and pattern recognition},
  pages={2027--2034},
  year={2013}
}

@article{landrieu2017cut,
  title={Cut pursuit: Fast algorithms to learn piecewise constant functions on general weighted graphs},
  author={Landrieu, Loic and Obozinski, Guillaume},
  journal={SIAM Journal on Imaging Sciences},
  volume={10},
  number={4},
  pages={1724--1766},
  year={2017},
  publisher={SIAM}
}

@inproceedings{zhu2024spgroup3d,
  title={{SPGroup3D}: Superpoint Grouping Network for Indoor {3D} Object Detection},
  author={Zhu, Yun and Hui, Le and Shen, Yaqi and Xie, Jin},
  booktitle={Proceedings of the AAAI Conference on Artificial Intelligence},
  volume={38},
  pages={7811--7819},
  year={2024}
}

@inproceedings{cheng2021sspc,
  title={Sspc-net: Semi-supervised semantic {3D} point cloud segmentation network},
  author={Cheng, Mingmei and Hui, Le and Xie, Jin and Yang, Jian},
  booktitle={Proceedings of the AAAI conference on artificial intelligence},
  volume={35},
  pages={1140--1147},
  year={2021}
}

@inproceedings{qi2017pointnet,
  title={Pointnet: Deep learning on point sets for {3D} classification and segmentation},
  author={Qi, Charles R and Su, Hao and Mo, Kaichun and Guibas, Leonidas J},
  booktitle={Proceedings of the IEEE conference on computer vision and pattern recognition},
  pages={652--660},
  year={2017}
}

@article{vaswani2017attention,
  title={Attention is all you need},
  author={Vaswani, Ashish and Shazeer, Noam and Parmar, Niki and Uszkoreit, Jakob and Jones, Llion and Gomez, Aidan N and Kaiser, {\L}ukasz and Polosukhin, Illia},
  journal={Advances in neural information processing systems},
  volume={30},
  year={2017}
}

@inproceedings{nguyen2024open3dis,
  title={Open{3D}is: Open-vocabulary {3D} instance segmentation with 2d mask guidance},
  author={Nguyen, Phuc and Ngo, Tuan Duc and Kalogerakis, Evangelos and Gan, Chuang and Tran, Anh and Pham, Cuong and Nguyen, Khoi},
  booktitle={Proceedings of the IEEE/CVF Conference on Computer Vision and Pattern Recognition},
  pages={4018--4028},
  year={2024}
}

@article{li2024instancegaussian,
  title={InstanceGaussian: Appearance-Semantic Joint Gaussian Representation for 3D Instance-Level Perception},
  author={Li, Haijie and Wu, Yanmin and Meng, Jiarui and Gao, Qiankun and Zhang, Zhiyao and Wang, Ronggang and Zhang, Jian},
  journal={arXiv preprint arXiv:2411.19235},
  year={2024}
}

@inproceedings{kim2024garfield,
  title={Garfield: Group anything with radiance fields},
  author={Kim, Chung Min and Wu, Mingxuan and Kerr, Justin and Goldberg, Ken and Tancik, Matthew and Kanazawa, Angjoo},
  booktitle={Proceedings of the IEEE/CVF Conference on Computer Vision and Pattern Recognition},
  pages={21530--21539},
  year={2024}
}

@article{cheng2024occam,
  title={Occam's LGS: A Simple Approach for Language Gaussian Splatting},
  author={Cheng, Jiahuan and Zaech, Jan-Nico and Van Gool, Luc and Paudel, Danda Pani},
  journal={arXiv preprint arXiv:2412.01807},
  year={2024}
}

@inproceedings{drsplat25,
    title={Dr. Splat: Directly Referring 3D Gaussian Splatting via Direct Language Embedding Registration},
    author={Jun-Seong, Kim and Kim GeonU and Yu-Ji, Kim and Yu-Chiang Frank Wang and Jaesung Choe and Oh, Tae-Hyun},
    booktitle=CVPR,
    year={2025}
}

@article{goi2024,
    title={GOI: Find 3D Gaussians of Interest with an Optimizable Open-vocabulary Semantic-space Hyperplane},
    author={Qu, Yansong and Dai, Shaohui and Li, Xinyang and Lin, Jianghang and Cao, Liujuan and Zhang, Shengchuan and Ji, Rongrong},
    journal={arXiv preprint arXiv:2405.17596},
    year={2024}
}

@article{engelmann2024opennerf,
  title={OpenNeRF: open set 3D neural scene segmentation with pixel-wise features and rendered novel views},
  author={Engelmann, Francis and Manhardt, Fabian and Niemeyer, Michael and Tateno, Keisuke and Pollefeys, Marc and Tombari, Federico},
  journal={arXiv preprint arXiv:2404.03650},
  year={2024}
}

@article{chen2024panoptic,
  title={Panoptic vision-language feature fields},
  author={Chen, Haoran and Blomqvist, Kenneth and Milano, Francesco and Siegwart, Roland},
  journal={IEEE Robotics and Automation Letters},
  volume={9},
  number={3},
  pages={2144--2151},
  year={2024},
  publisher={IEEE}
}

@article{qi2017pointnet++,
  title={Pointnet++: Deep hierarchical feature learning on point sets in a metric space},
  author={Qi, Charles Ruizhongtai and Yi, Li and Su, Hao and Guibas, Leonidas J},
  journal={Advances in neural information processing systems},
  volume={30},
  year={2017}
}

@article{silva2024contrastive,
  title={Contrastive gaussian clustering: Weakly supervised 3d scene segmentation},
  author={Silva, Myrna C and Dahaghin, Mahtab and Toso, Matteo and Del Bue, Alessio},
  journal={arXiv preprint arXiv:2404.12784},
  year={2024}
}
}
\clearpage
\setcounter{page}{1}
\maketitlesupplementary

\appendix

This supplementary document provides additional details about our method. \sectionautorefname~\ref{sec:superg_details} elaborates on the design of the \acrlong{superg} and demonstrates its application to downstream tasks. \sectionautorefname~\ref{sec:more_impl} provides further implementation details, including the MLP architectures for neural Gaussian feature decoding and the adaptation of OpenGaussian for 2D open-vocabulary semantic segmentation comparison. \sectionautorefname~\ref{sec:efficiency} reports detailed efficiency analysis, including training time, inference speed, and memory consumption. \sectionautorefname~\ref{sec:more_abl} reports extended ablation studies on \acrshort{superg} hyperparameters and module variants. \sectionautorefname~\ref{sec:more_result} presents additional quantitative and qualitative results. Lastly, \sectionautorefname~\ref{sec:limitation} discusses the limitations of our approach and outlines potential directions for future work.

\section{\acrlong{superg} Details}
\label{sec:superg_details}

\begin{figure}[htbp]
    \centering 
    \includegraphics[width=0.8\linewidth]{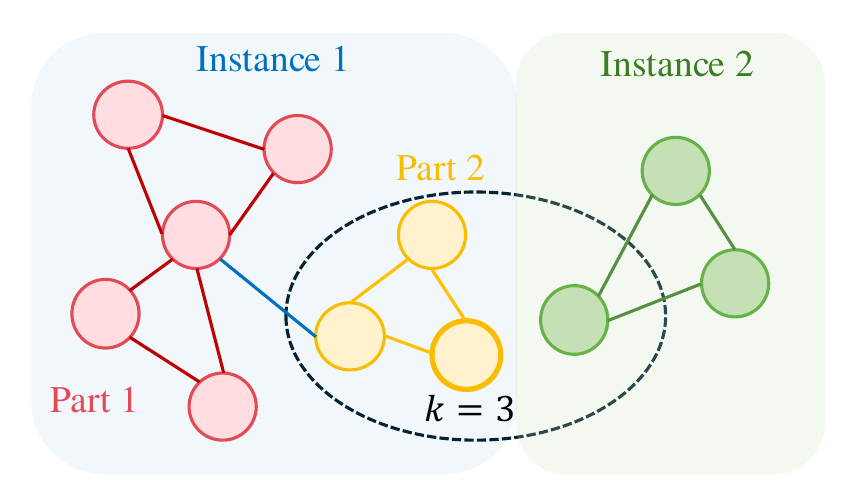} 
    \caption{Example of \acrlong{superg} graph. Each node represents a \acrshort{superg}, connected to its $k$-nearest neighbors based on similarity in instance features. Through connected component analysis, the SuperG nodes are divided into two distinct instances. Within Instance 1, SuperG nodes are interconnected by similar hierarchical features, further splitting into two parts.}
    \label{fig:sg_graph}
\end{figure}

\paragraph{Module Design.}
As shown in~\figureautorefname~\ref{fig:sg_module}, our \acrshort{superg} clustering network consists of four learnable MLPs. Inspired by SPNet~\cite{hui2021spnet}, we design three attribute-specific learnable functions $F_\phi$, $F_\varphi$, and $F_\psi$, each implemented as a single-hidden-layer MLP with ReLU activation. These functions independently encode the differences between an anchor and its $k$-nearest \acrshort{superg}s in coordinates, segmentation features, and geometry, respectively, producing embeddings that reflect the relevance of each attribute for \acrshort{superg} assignment. A final MLP, $F_{sg}$, then concatenates these embeddings and integrates spatial, semantic, and geometric cues into a probabilistic assignment.

\paragraph{Grouping \acrlong{superg}s for Instance and Hierarchical Segmentation.} After training the \acrshort{superg} association modules, we obtain a soft association map $\mA \in \mathbb{R}^{N' \times k}$.
During inference, each anchor point is assigned to one of its $k$-nearest neighbors with the highest probability, leading to a hard \acrshort{superg} assignment $\bar{\mA} \in \mathbb{R}^{N' \times 1}$.
The attributes of each \acrshort{superg} are then computed by averaging the attributes of its assigned anchors.

These \acrshort{superg}s serve as the fundamental units for representing and interpreting the 3D scene. Specifically, as shown in \figureautorefname~\ref{fig:sg_graph}, we further construct a graph where nodes correspond to \acrshort{superg}s. For instance segmentation, a node is connected to nodes within its $k$-nearest neighbors if their instance feature similarity exceeds a threshold $\tau_{Ins}$. Instances are then obtained via connected component analysis on this restricted graph. Similarly, part segmentation is achieved by building a \acrshort{superg} graph within each instance and identifying connected components based on hierarchical feature similarity with threshold $\tau_{Hier}$. In practice, we set $k=3$, $\tau_{Ins}=0.8$, and $\tau_{Hier}=0.9$.

\section{Additional Implementation Details}
\label{sec:more_impl}

\begin{figure}[htbp]
    \centering 
    \includegraphics[width=\linewidth]{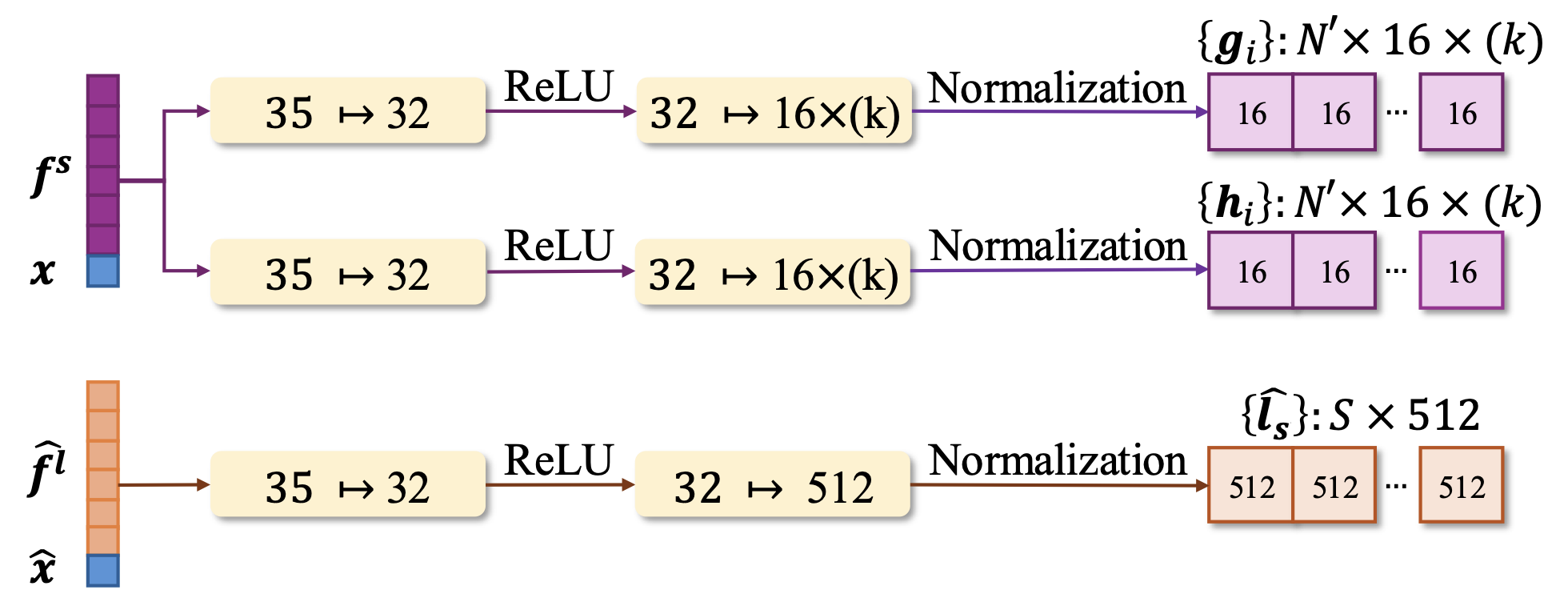} 
    \caption{MLP structures for decoding different features.}
    \vspace{-5mm}
    \label{fig:mlp_decoder}
\end{figure}

\paragraph{Decoding Neural Gaussians from MLPs.}
We employ MLPs to decode latent features, as shown in \figureautorefname~\ref{fig:mlp_decoder}. Each MLP contains a single hidden layer of dimension 32. Their decoding targets, however, differ: instance feature decoder $F_I$ and hierarchical feature decoder $F_H$ decode anchor-level features, while language feature decoder $F_L$ decodes \acrshort{superg}-level features.
Specifically, $F_I$ and $F_H$ take the anchor segmentation feature $\vf^s$ and anchor position $\vx$ as input, and predict the instance feature $\vg$ and hierarchical feature $\vh$ of the neural Gaussians spawned per anchor. In contrast, $F_L$ predicts the CLIP-aligned feature $\widehat{\vl}$ for each \acrshort{superg}, conditioned on its latent language feature $\widehat{\vf}^l$ and its center $\widehat{\vx}$.

\paragraph{Additional Information on the Contrastive Losses.}

The contrastive instance feature loss $\mathcal{L}_{Ins}$, which is computed on the set of SAM-generated instance-level masks $\mathcal{M}$ and the rendered 2D instance feature map $\hat{\mG}$, has been discussed in \sectionautorefname~\ref{sec:feats}. Inspired by~\cite{ying2024omniseg3d}, we elaborate a similar yet more complicated contrastive hierarchical feature loss $\mathcal{L}_{Hier}$. This loss is defined on the part-level patches $\mathcal{P} = \{{\vp}^p \in \sR^{H \times W} \mid p=1, \dots, |\mathcal{P}|\}$ and the rendered 2D hierarchical feature map $\hat{\mH} \in \sR^{D_h \times H \times W}$.

Given a patch $\vp^p$, we collect the rendered hierarchical features from $\hat{\mH}$ at each pixel, forming a set of feature vectors ${\vh^p}$. The mean feature of the patch is then defined as $\bar{\vh}^p$. We define the contrastive hierarchical feature loss at the minimum unit, on a pixel $t$ with feature $\vh^p_t \in {\vh^p}$, as:

\begin{equation}
\mathcal{L}^{p,t}(r) = -\log \frac{\exp(\vh^p_t \cdot \bar{\vh}^r / \tau_r)}{\sum_{q=1}^{|\mathcal{P}|} \exp(\vh^p_t \cdot \bar{\vh}^q / \tau_q)},
\end{equation}
where $\tau$ is the temperature of the contrastive loss, and $p, r, q$ are indices of patches.
Subsequently, the hierarchical feature loss~\cite{ying2024omniseg3d} can be written as:
\begin{equation}
\mathcal{L}_{Hier} = \sum_{p=1}^{|\mathcal{P}|} \sum_{d=1}^{d^p_{\text{max}}} \mathcal{L}_{p,d},
\vspace{-3mm}
\end{equation}

\begin{equation}
\mathcal{L}_{p,d} = \frac{\lambda^{d-1}}{|\mathcal{R}^p_d|} \sum_{t=1}^{|\{\vh^p\}|}\sum_{r \in \mathcal{R}_d^p} \max \left( \mathcal{L}^{p,t}(r), \mathcal{L}^{p,t}_{\max}(d-1) \right ),
\end{equation}
where $\lambda^{d-1}$ is a hyperparameter, $\mathcal{R}^p_d$ denotes the index set of patches at hierarchy level $d$ of patch $p$, and $r \in \mathcal{R}_{d}^p$ refers to a patch at level $d$. The maximum loss at level $d$ ensures that the contrastive loss between the pixel feature $t$ and patches with higher correlation (lower $d$) is always smaller than for patches with lower correlation: 

\begin{equation}
    \mathcal{L}^{p, t}_{max}(d) = \max_{r \in \mathcal{R}_d^p} \mathcal{L}^{p, t}(r).
    \vspace{-2mm}
\end{equation}

\paragraph{Additional Technical Details.} We use the SAM ViT-H model~\cite{kirillov2023sam} to generate 2D masks from the input images and then extract language features for each instance mask using the OpenCLIP ViT-B/16 model following~\cite{qin2024langsplat}. The training process is divided into three stages. In the first stage, we train the Scaffold-GS~\cite{lu2024scaffold} with instance and hierarchical feature attributes for 30k iterations. In the second stage, we freeze the geometry and multi-granularity features network from stage one and train only the \acrshort{superg} clustering network for another 30k iterations. Finally, in the last stage, we freeze all other parameters and optimize the language features for each SuperG for 10k iterations. The embedding dimensions for $\vf^g$ and $\vf^s$ are set to 32~\cite{lu2024scaffold}, while instance and hierarchical features are 16-dimensional~\cite{ying2024omniseg3d}. For optimization, we use the Adam~\cite{KingBa15} optimizer for the MLPs with an initial learning rate of 0.01 and an exponential annealing schedule of 0.001 as in~\cite {fridovich2022plenoxels}.

\paragraph{OpenGaussian Implementation.} OpenGaussian~\cite{wu2024opengaussian} assigns language features to instance-level Gaussians, enabling direct language queries on 3D point clouds. However, this approach does not natively support 2D pixel-level semantic segmentation, making direct evaluation on ScanNet more challenging.
To enable a fair comparison, we first identify category-relevant 3D Gaussians by iterating over all text prompts to predict language feature maps for open-vocabulary semantic segmentation. For each instance-level Gaussian cluster, we determine the corresponding text prompt ID and store these IDs in a label map, which is then used to generate the final semantic segmentation.
By following this approach, occlusions at the instance-level Gaussians are not explicitly handled, leading to the occlusion artifacts observed in \figureautorefname~\ref{fig:scannet_qualitative}.

\begin{table*}
\centering
\begin{tabular}{l|cc|cc|cc|cc|cc}
\toprule
& \multicolumn{2}{c|}{\textbf{mean}} & \multicolumn{2}{c|}{\textit{figurines}} & \multicolumn{2}{c|}{\textit{teatime}} & \multicolumn{2}{c|}{\textit{ramen}} & \multicolumn{2}{c}{\textit{waldo\_kitchen}} \\
Ablation & mIoU & mAcc & mIoU & mAcc & mIoU & mAcc & mIoU & mAcc & mIoU & mAcc \\
\midrule

$S = 250$ & 25.75 & 39.42 & 22.70 & 35.71 & 35.93 & 50.85 & 15.44 & 21.13 & 28.92 & \textbf{50.00} \\
$S = 500$ & \cellcolor{orange!20}34.05 & \cellcolor{orange!20}50.03 & 27.44 & 48.21 & \textbf{57.42} & \textbf{77.97} & \textbf{20.42} & \textbf{23.94} & \textbf{30.93} & \textbf{50.00} \\
$\mathbf{S = 1000}$& \cellcolor{red!20}\textbf{35.94} & \cellcolor{red!20}\textbf{52.02} & \textbf{43.68} & \textbf{60.71} & 55.31 & \textbf{77.97} & 18.07 & \textbf{23.94} & 26.71 & 45.45 \\
$S = 2000$& \cellcolor{yellow!20}27.61 & \cellcolor{yellow!20}40.60 & 27.64 & 46.43 & 47.58 & 62.71 & 12.46 & 16.90 & 22.76 & 36.36 \\
\midrule
$\mathbf{k = 3}$ & \cellcolor{red!20}\textbf{35.94} & \cellcolor{red!20}\textbf{52.02} & \textbf{43.68} & 60.71 & 55.31 & 77.97 & 18.07 & 23.94 & 26.71 & \textbf{45.45} \\
$k = 5$& \cellcolor{yellow!20}33.70 & \cellcolor{yellow!20}46.76 & 21.59 & 35.71 & \textbf{68.75} & \textbf{84.75} & 16.96 & 21.13 & 27.48 & \textbf{45.45} \\
$k = 10$& \cellcolor{orange!20}33.81 & \cellcolor{orange!20}46.88 & 43.56 & \textbf{62.50} & 41.56 & 55.93 & \textbf{21.82} & \textbf{28.17} & \textbf{28.31} & 40.91 \\
\bottomrule
\end{tabular}
\caption{Additional ablation studies on the LERF-OVS dataset~\cite{qin2024langsplat} about the parameter choices for the \acrshort{superg} Clustering Network. We use $s=1000$ \acrshort{superg}s and $k=3$ for $k$-nearest neighbor in our implementation by default.}
\label{tab:s_k}
\end{table*}

\section {Training and Inference Efficiency}
\label{sec:efficiency}
\begin{table}[htpb]
\setlength{\tabcolsep}{2.38pt}
	\centering
    \vspace{-3mm}
		\begin{tabular}{c|cc|ccc|ccc}
			\toprule
			\multirow{2}{*}{} & \multicolumn{2}{c|}{LangSplat} & \multicolumn{3}{c|}{OpenGaussian} & \multicolumn{3}{c}{Ours}  \\ 
                        &S1  & S2  & S1&S2 &S3 & S1&S2 &S3   \\ 
                        \midrule
			Train &  20m &  45m  & 50m & 20m & \textbf{10m} &  90m & 85m & 30m\\
			Memory    & \textbf{8G} & 6G &  14G & 17G & 22G & 18G & 14G & \textbf{4G}\\ 
                        \midrule
                Inference & \multicolumn{2}{c|}{3.28s} & \multicolumn{3}{c|}{5.55s}&\multicolumn{3}{c}
                {\textbf{0.56s}}\\
                Memory &\multicolumn{2}{c|}{18GB} & \multicolumn{3}{c|}{9GB}&\multicolumn{3}{c}
                {\textbf{4GB}}\\
			\bottomrule
		\end{tabular}
    \vspace{-3mm}
	\caption{Comparison of the time and GPU memory requirements during training (top rows) and inference (bottom rows).}
	\label{tab:efficiency}
\end{table}

In Table~\ref{tab:efficiency}, we report training time, inference time and memory consumption for LangSplat~\cite{qin2024langsplat} and OpenGaussian~\cite{wu2024opengaussian} on the LERF-OVS dataset.

For LangSplat~\cite{qin2024langsplat}, the training consists of two stages: in S1, the {3DGS} is pretrained without any additional feature fields for 30k iterations. In S2, the pretrained {3DGS} is frozen, and a language feature field is optimized for another 30k iterations.
For OpenGaussian~\cite{wu2024opengaussian}, S1 corresponds to pretraining the {3DGS} jointly with an instance feature field for 40k iterations. In S2, Gaussian clustering is performed in a coarse-to-fine manner, requiring an additional 30k iterations. Finally, in S3, the 2D language features are directly associated with the 3D Gaussian clusters.

When comparing our method to LangSplat and OpenGaussian across different training stages S1, S2, and S3, we find that our approach, while requiring longer training times in S1 and S2, achieves comparable efficiency in S3, indicating a trade-off due to the joint learning of instance and hierarchical features. GPU memory usage varies, with our method consuming more resources in S1 and S2 but significantly less in S3, showcasing the compact memory footprint enabled by our proposed \acrshort{superg}s. Conversely, LangSplat exhibits consistent memory efficiency in S1 and S2, while OpenGaussian's memory demands vary across different training stages. Most notably, our method consistently outperforms both baselines in inference speed across all scenarios, a critical advantage for real-time applications. This blend of rapid inference and flexible resource utilization highlights our method's robustness for practical deployment in diverse computer vision tasks. In addition, our method supports multi-granularity scene understanding. Specifically, it enables semantic-level queries to retrieve groups of objects sharing the same language description, instance-level segmentation of a specific object, and further decomposition of this instance into fine-grained parts. In contrast, LangSplat~\cite{qin2024langsplat} only supports semantic-level queries. OpenGaussian~\cite{wu2024opengaussian} extends to instance-level understanding but, similar to LangSplat, does not support finer-grained part segmentation. This demonstrates that our method provides a more comprehensive representation for 3D scene understanding.

\section{Additional Ablation Studies}

\label{sec:more_abl}

In \sectionautorefname~\ref{sec:experiment}, we perform ablation studies to evaluate the necessity of \acrshort{superg}s and analyze the performance of the instance feature field and hierarchical feature field. In this section, we further investigate the necessity of our proposed \acrshort{superg} clustering network and measure the impact of its individual components.

\paragraph{\acrlong{superg} Clustering Approaches.}

Given a pre-trained scene, our objective is to group anchors into meaningful \acrshort{superg}s using their coordinates, segmentation features, and geometric properties. We explore two alternative approaches. First, we evaluate a simple $K$-means clustering algorithm by concatenating the aforementioned attributes and clustering them into $k=1000$ \acrshort{superg}s. Second, we experiment with a traditional supervoxel generation approach. Specifically, we map the anchors to a point cloud, using the concatenated features as normals. We then apply the Voxel Cloud Connectivity Segmentation (VCCS) algorithm~\cite{papon2013voxel} to compute the \acrshort{superg}s. Finally, we compare these two non-learning-based approaches with our learning-based method for Anchor-to-\acrshort{superg} association.

We observed that $K$-means fails to prevent the overlap of resulting \acrshort{superg}s across instances. Meanwhile, VCCS~\cite{papon2013voxel}, originally designed for dense point clouds, struggles with the sparse structure of Gaussians. Its region-growing mechanism incorrectly clusters a large number of anchors together, which hinders the learning of the language feature field. The results in \tableautorefname~\ref{table:abl_sg_method} show that our method is better suited for grouping Gaussians, achieving better performance.

\begin{table}[thb]
\setlength{\tabcolsep}{3pt}
\vspace{-0.2em}
  \centering
  \begin{tabular}{c|cc}
    \toprule
    Method & mIoU $\uparrow$ & mAcc. $\uparrow$ \\
    \midrule
     $K$-means & 53.77 & 67.80 \\
     VCCS\cite{papon2013voxel} & 0.45 & 0.00 \\
     Ours & \textbf{55.31} & \textbf{77.97} \\
    \bottomrule
  \end{tabular}
  \vspace{-0.4em}
  \caption{Ablation study of \acrshort{superg} clustering approaches on the \textit{teatime} scene of LERF-OVS.}
  \label{table:abl_sg_method}
  \vspace{-0.2em}
\end{table}

\begin{table}[thb]
\setlength{\tabcolsep}{3pt}
\vspace{-0.2em}
  \centering
  \begin{tabular}{ccc|cc}
    \toprule
    w/ $F_\phi$ & w/ $F_\varphi$ & w/ $F_\psi$ & mIoU $\uparrow$ & mAcc. $\uparrow$ \\
    \midrule
     \ & \ & \ & 32.41 & 40.68 \\
     \checkmark & \ & \ & 48.29 & 67.80 \\
     \ & \checkmark & \ & \textbf{58.07} & 75.66 \\
     \ & \ & \checkmark & 37.12 & 62.71 \\
     \checkmark & \checkmark & \checkmark & 55.31 & \textbf{77.97} \\
    \bottomrule
  \end{tabular}
  \vspace{-0.4em}
  \caption{Ablation study on the \acrshort{superg} clustering network and its components on the \textit{teatime} scene of LERF-OVS.}
  \label{table:abl_sg_comp}
  \vspace{-0.2em}
\end{table}

\begin{table*}[ht]
\setlength{\tabcolsep}{2.68pt}
\centering
\begin{tabular}{l|cc|cc|cc|cc|cc|cc|cc}
\toprule
& mIoU & mAcc & mIoU & mAcc & mIoU & mAcc & mIoU & mAcc & mIoU & mAcc & mIoU & mAcc & mIoU & mAcc \\
Method & \multicolumn{2}{c|}{\textbf{mean}} & \multicolumn{2}{c|}{\textit{wall}} & \multicolumn{2}{c|}{\textit{floor}} & \multicolumn{2}{c|}{\textit{cabinet}} & \multicolumn{2}{c|}{\textit{table}} & \multicolumn{2}{c|}{\textit{desk}} & \multicolumn{2}{c}{\textit{curtain}} \\
\midrule
LERF \cite{kerr2023lerf} &\cellcolor{orange!20}38.5 & \cellcolor{yellow!20}60.4& 35.2& 82.8& 60.1& 68.8& 52.0&82.7& 10.2& 80.1&  14.4& 16.1& 70.2& 77.8\\
LEGaussians \cite{shi2024language}    & 8.7  & 33.2 & 17.9 & 53.1 & 14.6 & 20.6 & 2.7  & 18.6 & 0.0  & 0.0 & 0.5 & 13.5  & 1.9  & 10.4 \\
OpenGaussian \cite{wu2024opengaussian}& \cellcolor{yellow!20}24.1 & \cellcolor{orange!20}68.7 & 13.4 & \textbf{96.6} & 31.2 & 74.4 & 0.3  & 22.9 & 0.1  & 1.0  & \textbf{30.6} & \textbf{35.6} & 17.7 & \textbf{79.2} \\
LangSplat \cite{qin2024langsplat}      & 27.6 & 48.3 & 45.3 & 72.6 & 43.3 & 45.6 & 24.8 & 56.7 & 21.9 & \textbf{87.4} & 0.1 & 6.4 & 46.8 & 66.5 \\
\textbf{SuperGSeg} [ours]      & \cellcolor{red!20}\textbf{54.7} & \cellcolor{red!20}\textbf{74.7} & \textbf{58.8} & 92.9 & \textbf{53.6} & \textbf{86.5} & \textbf{69.8} & \textbf{83.8} & \textbf{35.7} & 54.8 & 15.0 & \textbf{16.7} & \textbf{61.8} & 64.5 \\
\midrule
& \multicolumn{2}{c|}{\textit{toilet}} & \multicolumn{2}{c|}{\textit{counter}} & \multicolumn{2}{c|}{\textit{refrigerator}} & \multicolumn{2}{c|}{\textit{chair}} & \multicolumn{2}{c|}{\textit{sink}} & \multicolumn{2}{c|}{\textit{window}} & \multicolumn{2}{c}{\textit{door}} \\
\midrule
LERF \cite{kerr2023lerf}      & 25.2 & 25.2 & 24.4 & 42.8 & 69.9 & 90.2 & 10.9 & 10.9 & 25.8 & 37.1 & 11.5 & 11.5 & 64.5 & 67.5 \\
LEGaussians \cite{shi2024language}     & 13.7 & 16.3 & 10.7 & 27.0 & 9.0  & 74.3 & 0.4  & 28.7 & 0.3  & 0.4  & 0.0  & 44.4 & 1.4  & 4.7  \\
OpenGaussian \cite{wu2024opengaussian} & \textbf{73.0} & \textbf{98.4} & 3.0  & 9.3  & \textbf{88.0} & \textbf{98.3} & 36.5 & 83.4 & 3.0  & 3.7  & \textbf{75.0} & \textbf{88.8} & \textbf{75.4} & \textbf{97.0} \\
LangSplat \cite{qin2024langsplat}      & 0.1  & 5.4  & 10.7 & 34.7 & 0.7  & 33.3 & 18.0 & 48.5 & 0.0  & 0.0  & 0.0  & 0.1  & 55.6 & 66.3 \\
\textbf{SuperGSeg} [ours]      & 26.9 & 26.9 & \textbf{14.0} & \textbf{59.1} & 79.4 & 80.2 & \textbf{80.4} & \textbf{83.8} & \textbf{11.7} & \textbf{12.0} & 54.7 & 77.0 & 58.2 & 58.3 \\
\bottomrule
\end{tabular}
\vspace{-3mm}
\caption{Comparison of mIoU and mAcc for various methods on each class of the ScanNet v2 dataset~\cite{dai2017scannet}.}
\label{tab:scannet_more}
\vspace{-2mm}
\end{table*}

\paragraph{\acrlong{superg} Clustering Network.} As introduced in \sectionautorefname~\ref{sec:superg_details}, we employ three MLPs $F_\phi$, $F_\varphi$, and $F_\psi$ to capture the coordinate, segmentation, and geometric relationships between anchors and their $k$-nearest neighbors. To evaluate the contributions of these MLPs, we conduct an ablation study. Notably, in experiments where none of these MLPs are used, we directly concatenate the differences of the attributes as input to $F_{sg}$ for predicting the association matrix. The results presented in \tableautorefname~\ref{table:abl_sg_comp} demonstrate that each MLP contributes to improving the \acrshort{superg} assignments. The MLP $F_\varphi$, which accounts for the segmentation feature differences between the anchor and the \acrshort{superg}, has the most significant impact. In particular, using only $F_\varphi$ yields a relatively high mIoU, emphasizing its effectiveness in aligning semantic features. However, our full setup that integrates the $F_\phi$ and $F_\psi$ for coordinate and geometric feature information further enhances mAcc. This suggests that incorporating additional spatial and geometric context refines the \acrshort{superg} assignments, leading to a more precise understanding of the scene.

\paragraph{Parameters in \acrlong{superg} Clustering Network.} We conduct ablation studies on the parameters involved in generating \acrshort{superg}s using the \acrshort{superg} clustering network. One crucial parameter is the total number of \acrshort{superg}s predefined, denoted as $S$. Too few \acrshort{superg}s fail to distinguish all instances, causing a single \acrshort{superg} to span multiple instances, which undermines semantic accuracy. Conversely, too many \acrshort{superg}s may introduce additional noise. Another parameter is the number of neighboring \acrshort{superg}s $k$ considered for each anchor when computing the association matrix between anchors and \acrshort{superg}s.

As shown in \tableautorefname~\ref{tab:s_k}, these two parameters are highly scene-specific, with the optimal number of \acrshort{superg}s $S$ and neighbors $k$ varying across different scenes. Notably, for fair comparisons, we use the same parameter values, $S=1000$ and $k=3$, for all scenes. This parameter choice achieves optimal performance on average.

\section{Additional Results}
\label{sec:more_result}

\begin{figure}[htbp]
    \centering
    \begin{subfigure}{0.24\linewidth}
        \includegraphics[width=\linewidth]{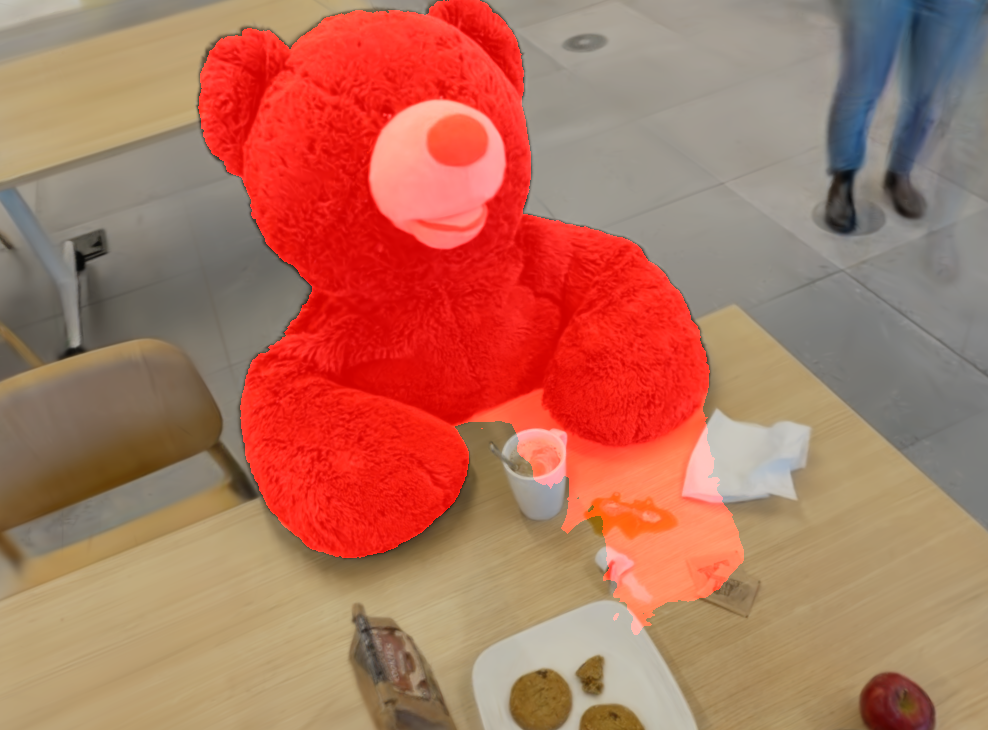}
    \end{subfigure}
    \begin{subfigure}{0.24\linewidth}
        \includegraphics[width=\linewidth]{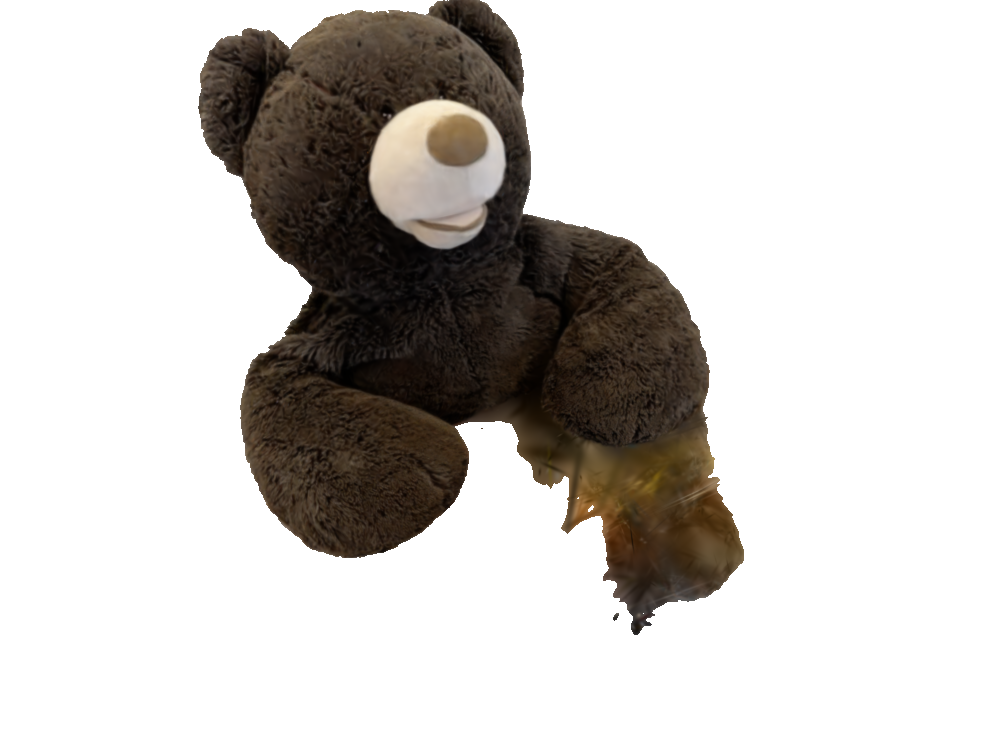}
    \end{subfigure}
    \begin{subfigure}{0.24\linewidth}
        \includegraphics[width=\linewidth]{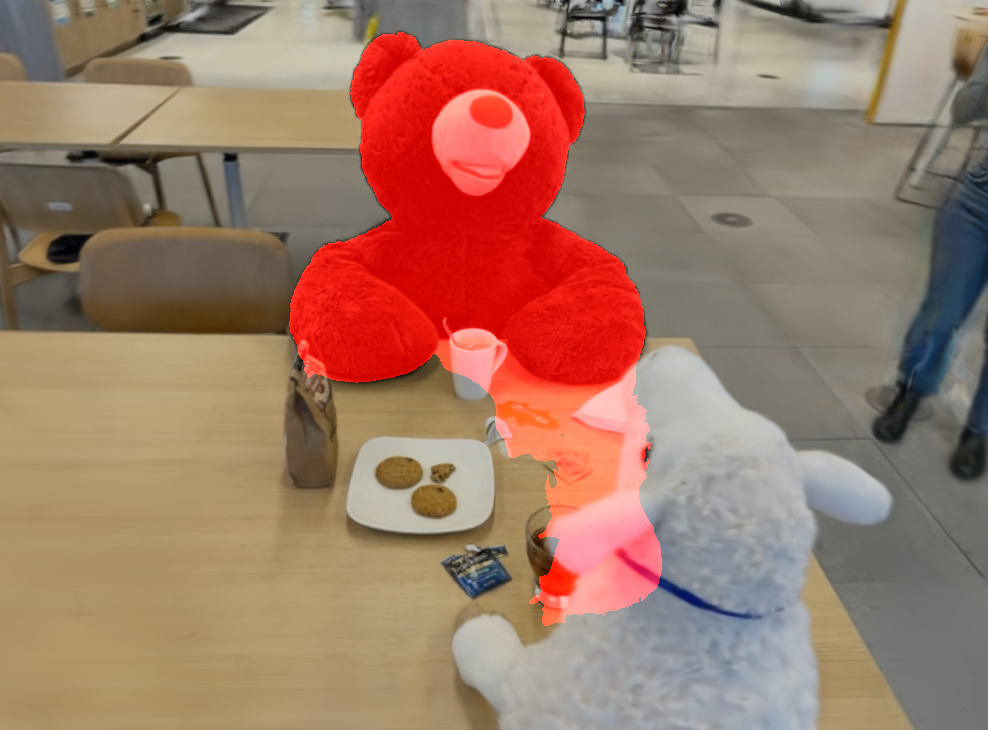}
    \end{subfigure}
    \begin{subfigure}{0.24\linewidth}
        \includegraphics[width=\linewidth]{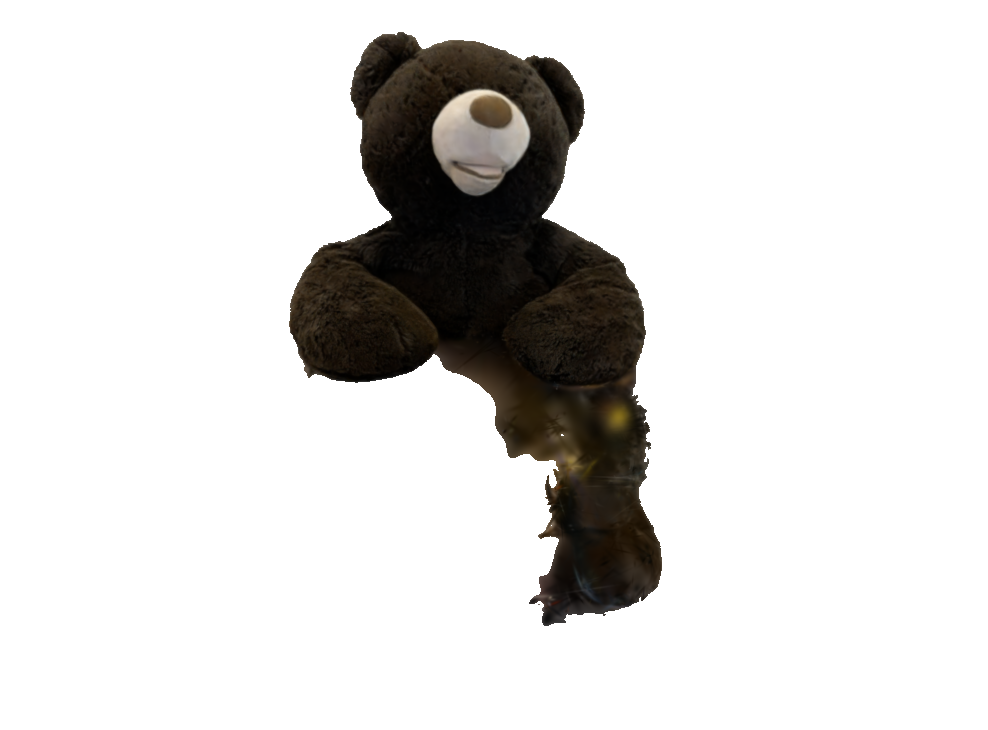}
    \end{subfigure}
    \begin{subfigure}{0.24\linewidth}
        \includegraphics[width=\linewidth]{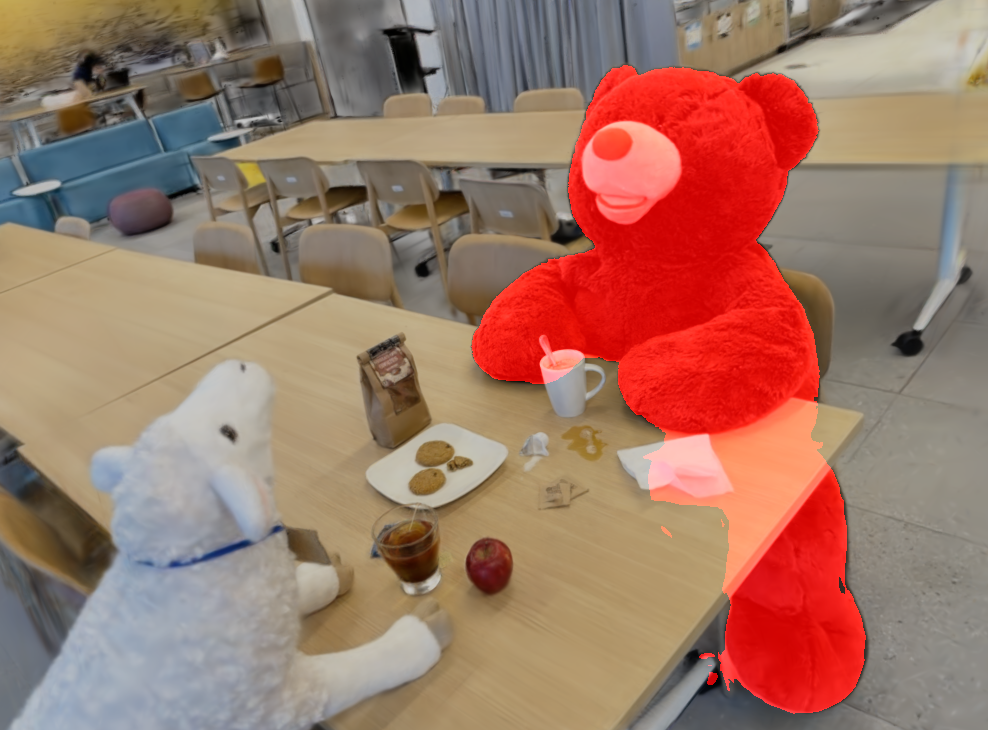}
    \end{subfigure}
    \begin{subfigure}{0.24\linewidth}
        \includegraphics[width=\linewidth]{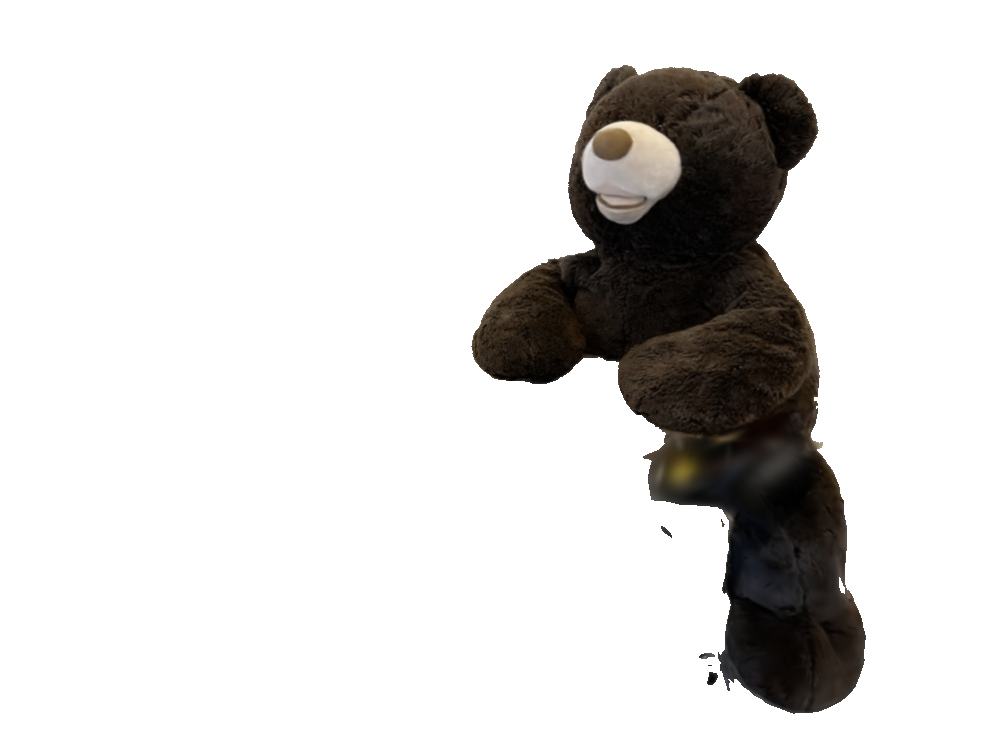}
    \end{subfigure}
    \begin{subfigure}{0.24\linewidth}
        \includegraphics[width=\linewidth]{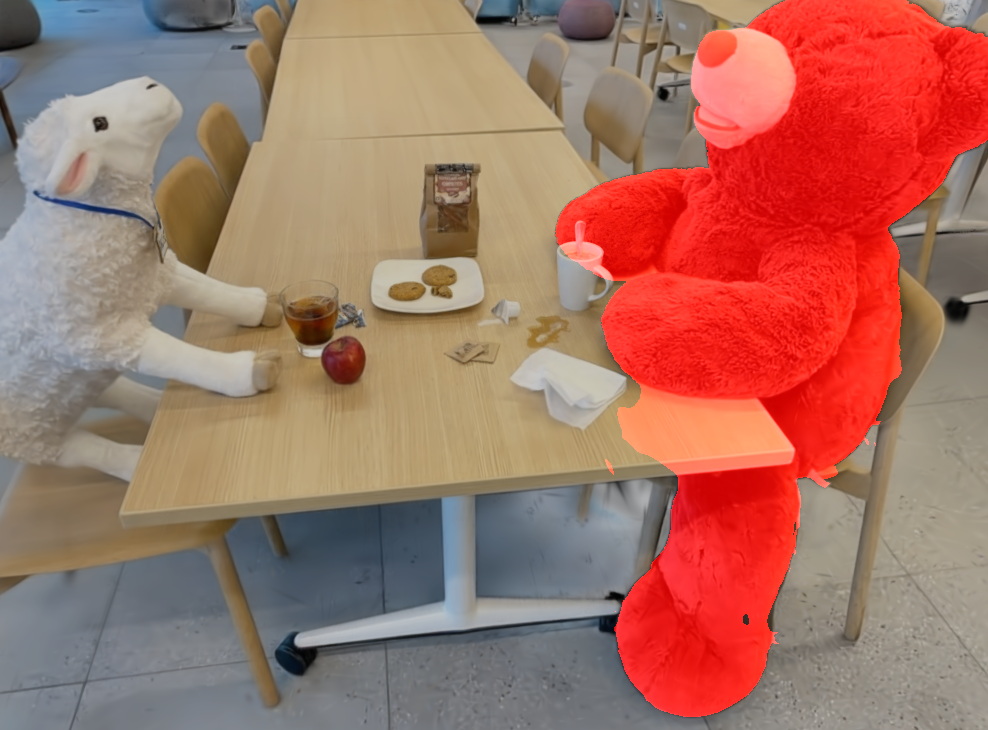}
    \end{subfigure}
    \begin{subfigure}{0.24\linewidth}
        \includegraphics[width=\linewidth]{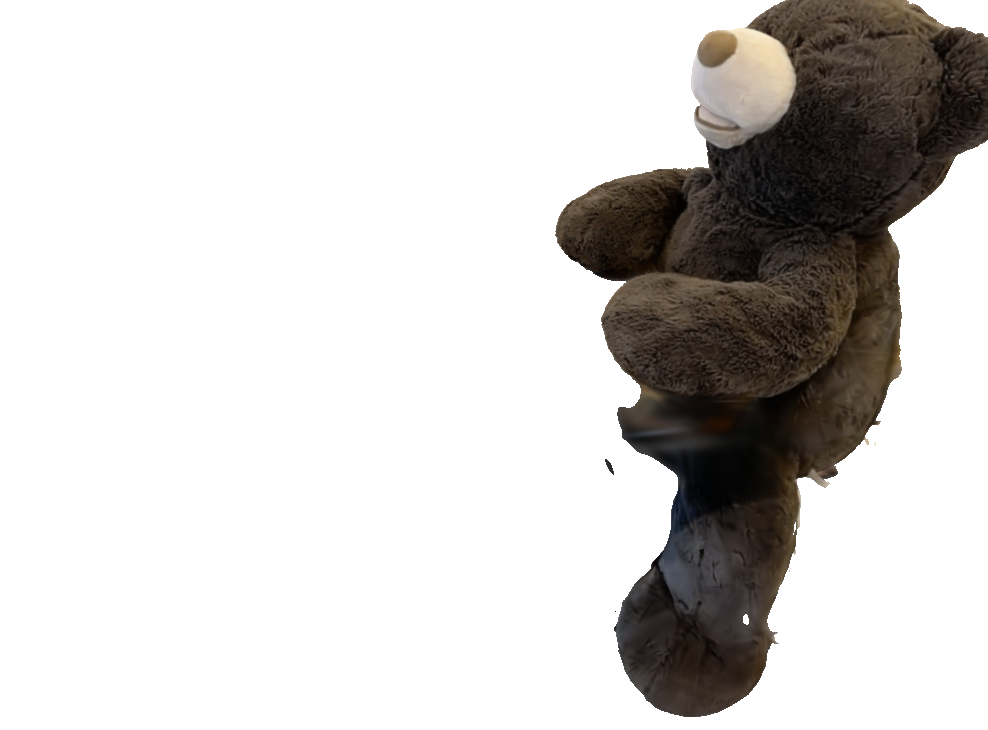}
    \end{subfigure}
    \caption{The language-queried 3D masks rendering to arbitrary viewpoints remain multi-view consistent. Benefit from the 3D understanding, we enable render regions that are originally occluded and invisible in 2D.}
    \label{fig:horizontal_subfigures}
\end{figure}

\textbf{Qualitative Results in Occlusion Cases.}
As illustrated in~\figureautorefname~\ref{fig:horizontal_subfigures}, our method queries objects directly in 3D space, effectively mitigating occlusion issues (e.g., the bear leg under the table can be retrieved). Moreover, the queried objects exhibit multi-view consistency, enabling comprehensive scene understanding in 3D. For additional qualitative results, please refer to the accompanying videos.

\paragraph{Additional Quantitative Results on ScanNet.} We report the results on more categories in the ScanNet dataset in~\tableautorefname~\ref{tab:scannet_more}.

\section{Limitations}
\label{sec:limitation}
Despite the advancements achieved by our method, certain limitations remain. First, our approach inherits biases from the original visual foundation models, which may constrain performance and limit generalization to diverse or unseen scenarios. Second, our method is tailored for scene-specific language representation, requiring significant modeling time for each scene. This limits its applicability in tasks that demand rapid adaptation or broad generalization, such as in-the-wild scene understanding. Future work could focus on mitigating inherited biases and optimizing training pipelines to enhance scalability and generalization.
\end{document}